\documentclass[10pt,conference]{IEEEtran}
\IEEEoverridecommandlockouts
\usepackage{cite}
\usepackage{amsmath,amssymb,amsfonts}
\usepackage{algorithmic}
\usepackage{graphicx}
\usepackage{textcomp}
\usepackage{xcolor}
\usepackage{booktabs}
\usepackage{multirow}
\usepackage{array}
\usepackage{url}
\usepackage[hidelinks]{hyperref}
\usepackage{tikz}
\usetikzlibrary{arrows.meta,positioning,fit}
\newcommand{\rcb}{ICAE-Bench}
\newcommand{\rcblite}{ICAE-Bench-Lite}
\def\BibTeX{{\rm B\kern-.05em{\sc i\kern-.025em b}\kern-.08em
    T\kern-.1667em\lower.7ex\hbox{E}\kern-.125emX}}

\usepackage{adjustbox}
\usepackage{amssymb}
\usepackage{xcolor}    
\usepackage[most]{tcolorbox}
\usepackage{cuted}
\usepackage{multicol}

\definecolor{promptboxblue}{RGB}{59, 130, 246}
\definecolor{promptboxgray}{RGB}{107, 114, 128}
\definecolor{promptboxlightgray}{RGB}{243, 244, 246}
\definecolor{promptboxgreen}{RGB}{34, 197, 94}
\newtcolorbox{promptbox}[1][]{
  enhanced,
  breakable,           
  colback=promptboxlightgray,
  colframe=promptboxblue!30,
  arc=8pt,
  boxrule=0.5pt,
  left=12pt,
  right=12pt,
  top=8pt,
  bottom=8pt,
  fonttitle=\bfseries,
  fontupper=\linespread{1.2}\selectfont,
  title=#1,
  before skip=10pt,
  after skip=10pt
}

\begin{document}

\title{ICAE-Bench: Evaluating Coding Agents as Interactive Project Builders}

\author{
\IEEEauthorblockN{
\begin{tabular}{c}
Zhongyuan Peng\textsuperscript{*,1,2},
Dan Huang\textsuperscript{*,3},
Chuyu Zhang\textsuperscript{2},
Caijun Xu\textsuperscript{1,4},
Changyi Xiao\textsuperscript{1}, 
Shibo Hong\textsuperscript{1} \\
David Lo\textsuperscript{3},
Lin Qiu\textsuperscript{2},
Xuezhi Cao\textsuperscript{2},
Jiyuan He\textsuperscript{\textdagger,2},
Yixin Cao\textsuperscript{\textdagger,1,4}
\end{tabular}
}
\IEEEauthorblockA{
\textsuperscript{1}Fudan University \quad
\textsuperscript{2}Meituan Group \quad
\textsuperscript{3}Singapore Management University \quad
\textsuperscript{4}Shanghai Innovation Institute
}
}


\IEEEtitleheader{%
  \vbox{%
    \hbox to \textwidth{%
      \includegraphics[height=0.4in]{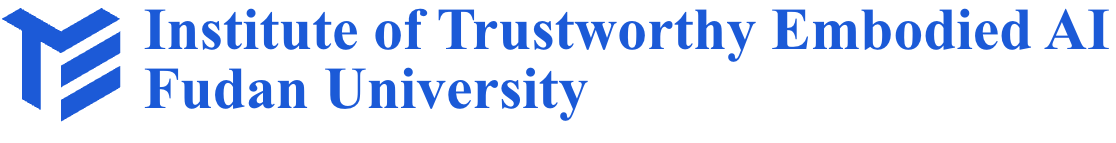}%
    }%
    \kern 1pt
    \hrule height 0.8pt width \textwidth
  }%
  \vskip -0.05em
}

\maketitle
\begingroup
\renewcommand{\thefootnote}{}
\footnotetext{\hspace*{-1.8em}\parbox[t]{0.95\columnwidth}{%
\textsuperscript{*}Equal contribution.\\
\textsuperscript{\textdagger}Corresponding authors: Jiyuan He (hejiyuan@meituan.com) and Yixin Cao (yxcao@fudan.edu.cn).}}
\endgroup

\begin{abstract}
The recent emergence of vibe-coding workflows is changing what coding agents are expected to do. Instead of merely completing code under fully specified instructions, agents are increasingly expected to transform incomplete product intent into working software by combining various abilities including planning, requirement clarification, tool use, debugging, and repository-level construction. Yet existing benchmarks have not fully caught up with this shift, evaluating agents on static, fully specified tasks.

In this paper, we introduce \rcb{}, an benchmark for evaluating coding agents under interactive project-building settings. The basic idea is to starts from a fuzzy product requirement, simulate the dynamic paradigm with an automated User Agent. To make this setting both realistic and evaluable, \rcb{} introduces three key designs. First, to avoid the ambiguity of unconstrained fuzzy requirements, each task derives ambiguity from precise real open-source repository with executable behavior. Second, to ensure high-quality and reproducible user simulation, \rcb{} grounds interaction through User Agent Data, allowing the User Agent to reveal hidden constraints without inventing new requirements or leaking implementation artifacts. Third, to evaluate open-ended repositories fairly, \rcb{} uses standardized black-box tests together with multi-dimensional diagnostics, including functional correctness, semantic and API similarity, structural fidelity, design quality, and interaction quality.

\rcb{} contains 480 tasks across 12 programming languages. Extensive experiments with six coding models and two agent frameworks show that ambiguous project generation remains challenging. Agents usually reproduce visible behavior but struggle with hidden constraints, boundary cases, and long-horizon integration. Our benchmark, dataset, and evaluation code are available at: \url{https://github.com/ALEX-nlp/ICAE-EVAL}

\end{abstract}


\section{Introduction}
\label{sec:introduction}

\begin{figure}[t]
\centering
\includegraphics[width=0.94\columnwidth]{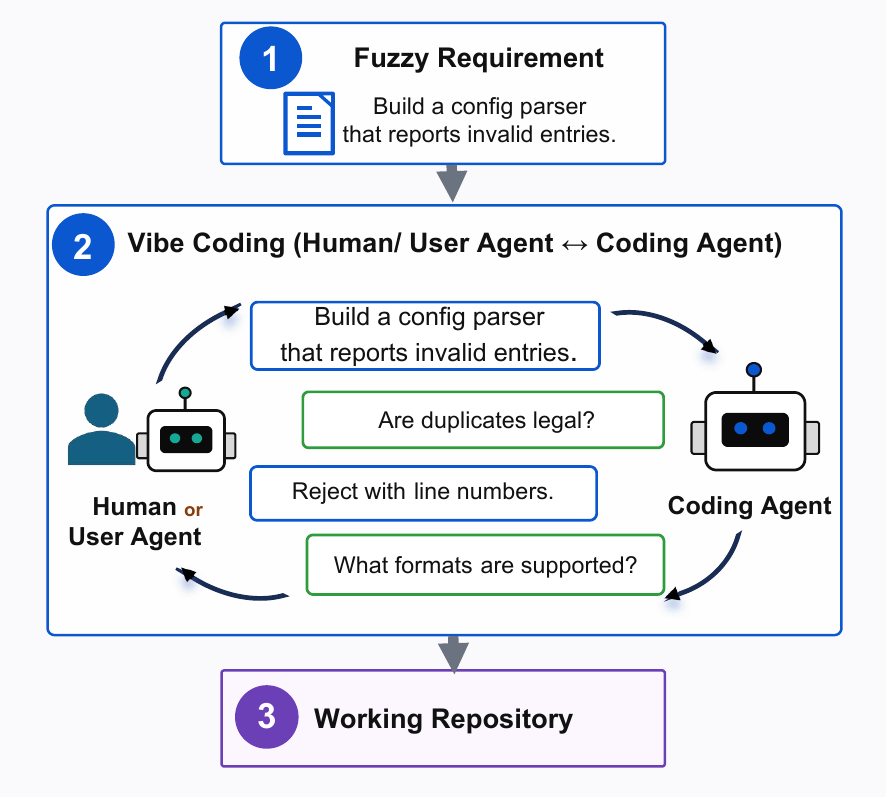}
\caption{\textbf{Overview of the interactive requirement clarification setting of \rcb{}.} Given an ambiguous requirement, vibe coding resolves missing design decisions through dialogue between a human and a coding agent. 
Our benchmark simulates this process with a user agent that answers clarification questions and guides the coding agent toward a complete, testable specification, which is then used to build a working repository.}
\label{fig:user-agent-interaction}
\end{figure}

Coding agents are shifting from code assistants to project builders. Early research largely centered on local tasks, such as completing a function or suggesting an edit.
By contrast, as shown in Figure~\ref{fig:user-agent-interaction}, recent ``vibe coding'' workflows often begin with a high-level user intent, instead of a complete and clear implementation contract. The coding agent must clarify missing requirements, inspect files, run commands, edit repositories, install dependencies, iterate over execution failures, and ultimately deliver a working project~\cite{yang2024sweagent,openhands}. This makes it increasingly plausible for users to delegate not only coding steps, but also larger parts of the software construction process. 
In this setting, the central question is whether a coding agent can turn incomplete human intent into functioning software.

This shift changes what coding benchmarks should evaluate, while existing ones still largely assume that the task goal is static, explicit, or discoverable without user interaction. Function-level benchmarks and repository-editing benchmarks measure localized synthesis, context use, and tool use inside existing codebases. From-scratch generation benchmarks move closer to project construction, but many still provide detailed requirements, fixed scaffolds, or language-specific constraints~\cite{zhao2024commit0,ding2026nl2repobench,fu2025prdbench}. 
A closely related step toward holistic software generation is ProgramBench~\cite{yang2026programbench}, which asks agents to rebuild programs from an executable and documentation. However, the target behavior is still exposed through a static interface. As a result, existing benchmarks remain limited in evaluating the broader capabilities required by vibe coding, where coding, planning, clarification, and iterative project construction are tightly coupled.

In this paper, we introduce \rcb{}, an \textbf{I}nteractive \textbf{C}oding \textbf{A}gents \textbf{E}valuation \textbf{Bench}mark for vibe-coding workflows.
In each task, a tested coding agent starts from a fuzzy requirement and may ask an automated User Agent for clarification. The User Agent answers from hidden requirement records instead of free-form speculation. Under this design, the central difficulty is not merely that the initial requirement is vague, but that vagueness poses three key challenges.

The first challenge is to ground ambiguity. In vibe-coding workflows, users rarely specify complete interfaces, edge cases, or architectural constraints upfront. However, if a benchmark directly writes vague requirements from scratch, the intended task may become underdefined. A failure may reflect an ambiguous or infeasible prompt rather than a limitation of the coding agent. We address this by deriving ambiguity from precision. Each task begins with a real open-source repository whose behavior is executable and testable. We first derive GroundPRD, a detailed product requirement document, together with black-box behavioral cases from the original repository, and then gradually fuzzify the requirement by hiding selected constraints. The hidden constraints are stored as User Agent Data, so the coding agent observes a fuzzy PRD while the benchmark retains a fixed behavioral target. This design preserves the realism of incomplete intent without losing the verifiability needed for evaluation.

To separate robustness to ambiguity from general implementation ability, we construct three fuzzy levels over the same task semantics. L1 provides the least initial information, L2 restores selected ambiguity points, and L3 retains most of GroundPRD while removing a limited set of details; GroundPRD is the fully specified endpoint. These levels form an operational information-exposure hierarchy rather than a universal psychometric scale. They enable controlled comparisons of how agents retrieve and use requirements as the initial specification becomes progressively richer.

The second challenge is to control interaction. A realistic project-building agent should be able to ask questions, but an uncontrolled user or free-form language model can introduce new requirements, leak implementation details, or hallucinate answers. \rcb{} therefore uses a grounded User Agent that can return only benchmark-authored requirement records. At the same time, the coding agent works in an ultimate image that preserves the runtime environment but removes golden code, original tests, and construction-time artifacts, preventing interaction and execution from becoming sources of benchmark leakage.

The third challenge is to evaluate open-ended repositories. Because project generation allows different internal designs, module structures, dependencies, APIs, and even implementation choices, a benchmark should not require the agent to reproduce the original repository literally. We constrain what the software must do, not how it must be implemented. Our multi-dimensional protocol combines black-box functional tests with artifact-, structure-, and interaction-level diagnostics to compare diverse implementations and localize breakdowns.

\rcb{} contains 480 tasks derived from real open-source repositories across 12 programming languages, together with a 50-task ICAE-Bench-Lite subset for efficient experimentation. Each task provides a fuzzy PRD, an ultimate image, User Agent Data through which omitted requirements and Public examples can be recovered, and authoritative black-box behavioral cases for evaluation. We evaluate six coding models within two agent frameworks, Claude Code and OpenHands. The experimental results suggest that i) GroundPRD remains a strong upper bound, interaction recovers only part of the gap, and ii) higher constraint coverage does not automatically translate into higher pass rate. The main bottleneck is not merely asking questions, but preserving clarified requirements and converting them into coherent repository implementations.


\begin{table}[t]
\centering
\caption{Comparison with representative from-scratch or repository-level coding benchmarks.}
\label{tab:benchmark-comparison}
\begin{adjustbox}{width=\linewidth, center} 
\begin{tabular}{lcccccc}
\toprule
\textbf{Benchmark} & \textbf{\#Proj.} & \textbf{Lang.} & \textbf{Req. Levels} & \textbf{Visible Info.} & \textbf{Interact.} & \textbf{Multi-PL Impl.} \\
\midrule
\multicolumn{7}{c}{\textbf{\textit{0-to-1 Generation Benchmarks}}} \\
\midrule
Commit0~\cite{zhao2024commit0} & 54 & 1 & Fixed & Impl.\textsuperscript{a} \& Test\textsuperscript{b} & $\times$ & $\times$ \\
NL2RepoBench~\cite{ding2026nl2repobench} & 104 & 1 & Fixed & Impl.\textsuperscript{a} & $\times$ & $\times$ \\
PRDBench~\cite{fu2025prdbench} & 50 & 1 & Fixed & Test\textsuperscript{b} & $\times$ & $\times$ \\
PRDBench(free)~\cite{fu2025prdbench} & 50 & 1 & Fixed & -- & $\times$ & $\times$ \\
ProgramBench~\cite{yang2026programbench} & 200 & 6 & Fixed & Prog.\textsuperscript{c} & $\times$ & $\times$ \\
RealBench~\cite{li2026realbench} & 61 & unknown & Fixed & Impl.\textsuperscript{a} & $\times$ & $\times$ \\
\midrule
\multicolumn{7}{c}{\textbf{\textit{Our Benchmarks}}} \\
\midrule
\textbf{\rcblite{}} & 50 & 10 & Fuzzy L1--L3\textsuperscript{d} & Pub. cases\textsuperscript{e} & \checkmark & \checkmark \\
\textbf{\rcb{}} & 480 & 12 & Fuzzy L1--L3\textsuperscript{d} & Pub. cases\textsuperscript{e} & \checkmark & \checkmark \\
\bottomrule
\end{tabular}
\end{adjustbox}

\vspace{3pt} 
\par\raggedright\scriptsize 
\textsuperscript{a} \textbf{Impl.}: Implementation details (e.g., internal APIs, structure, dependencies).\\
\textsuperscript{b} \textbf{Test}: Includes test scripts, test code, and all test data.\quad
\textsuperscript{c} \textbf{Prog.}: Executable programs.\\
\textsuperscript{d} L1--L3 are progressively richer Fuzzy PRDs; GroundPRD is the fully specified reference.\\
\textsuperscript{e} \textbf{Pub. cases}: Native examples exposed in GroundPRD and recoverable through interaction when omitted from Fuzzy L1/L2.
\end{table}


Our contributions are as follows:

\begin{itemize}
    \item We formulate interactive project generation under ambiguous product intent as a realistic benchmark setting for coding agents.
    \item We introduce a benchmark-construction pipeline and build \rcb{}, a 480-task benchmark spanning 12 languages, with language-agnostic black-box cases and controlled ultimate images.
    \item We conduct extensive experiments and provide a multi-dimensional evaluation covering functional correctness, artifact quality, structural fidelity, interaction quality, and controlled factors.
\end{itemize}

\section{Related Work}
\label{sec:related-work}

\subsection{Software-Agent Benchmarks}

Code-agent evaluation has gradually moved from isolated function synthesis toward repository-level and tool-using software engineering. Function-level benchmarks such as HumanEval~\cite{chen2021codex} and MBPP~\cite{austin2021program} measure local programming ability under relatively self-contained problem statements~\cite{chen2021evaluating,austin2021programsynthesislargelanguage}. 
Later benchmarks extend the evaluation context from single functions to files, repositories, and executable development environments. 
RepoBench~\cite{liu2023repobench} evaluates repository-level code completion with cross-file retrieval and completion tasks, while M2RC-Eval~\cite{liu2025m2rc} studies multi-file and multi-repository contextual reasoning. 
Issue-resolution benchmarks further place agents inside existing repositories and ask them to generate patches for real software issues, including SWE-bench~\cite{jimenez2024swebench}, Multi-SWE-bench~\cite{zan2025multi}, SWE-Bench Pro~\cite{deng2025swe}, and SWE-Compass~\cite{xu2025swe}. 
Recent surveys summarize this broader transition from code models to autonomous software-engineering agents~\cite{llm_codegen_survey2024,llm_agents_se_survey2024,codegen_agents_survey2025}.
These benchmarks are essential for measuring local coding, repository context use, tool execution, and issue-driven editing. However, they mostly assume that the target software context already exists and that the agent`s goal is to complete, edit, or repair it. They therefore do not directly evaluate the emerging project-building setting, which is main focus of the following works as well as ours.

\subsection{0-to-1 Repository Generation}

Existing 0-to-1 repository benchmarks differ mainly in how much of the target project is visible to the agent before implementation begins. One common design is to preserve substantial implementation scaffolding while removing selected code only. Commit0~\cite{zhao2024commit0}, for example, keeps repository structure, class definitions, and function signatures, so the core challenge is to fill in missing implementations rather than clarify underspecified product requirements. A second design provides a full natural-language project specification and asks the agent to realize it in an empty workspace. NL2RepoBench~\cite{ding2026nl2repobench}, RepoGenesis~\cite{peng2026repogenesisbenchmarkingendtoendmicroservice}, PRDBench~\cite{fu2025prdbench}, ProjectEval~\cite{liu2025projecteval}, and RealBench~\cite{li2026realbench} all move closer to end-to-end project building, but they still largely assume that the agent has been given the decisive requirements needed to start implementation.

Another line of work relaxes implementation constraints while keeping the behavioral target available through static artifacts. ProgramBench~\cite{yang2026programbench} evaluates whether agents can reconstruct behaviorally equivalent programs from executables and documentation, emphasizing black-box behavioral fidelity. This is closer to our language-agnostic evaluation philosophy, but its target behavior is still directly specified rather than partially hidden behind an interactive clarification process.

Table~\ref{tab:benchmark-comparison} summarizes these differences along the dimensions most relevant to interactive project generation: project scale, language coverage, initially visible information, clarification interaction, and cross-language evaluation. Compared with existing 0-to-1 benchmarks, \rcb{} exposes less decisive information at the start, while allowing hidden requirements to be recovered through multi-turn interaction. It also evaluates generated repositories with language-agnostic black-box tests, enabling cross-language implementations. Thus, \rcb{} complements prior 0-to-1 benchmarks by moving the focus from implementing a given specification to discovering and implementing a partially specified product target.


\subsection{Ambiguity and Clarification}

A growing line of work studies how code models handle unclear, incomplete, or interactive specifications~\cite{jin2026reqelicitgymevaluationenvironmentinterview,li2023pythoncodegenerationasking,Fakhoury_2024}. 
HumanEvalComm modifies HumanEval-style problems with inconsistency, ambiguity, and incompleteness to evaluate whether models ask useful clarification questions~\cite{Wu_2025}. 
Orchid studies the impact of requirement ambiguity on function-level code generation and categorizes ambiguity into lexical, syntactic, semantic, and vagueness-related forms~\cite{yang2026assessingimpactrequirementambiguity}. 
When Benchmarks Talk converts static code benchmarks into interactive settings where models must obtain key missing information from a simulated user~\cite{pan2025benchmarkstalkreevaluatingcode}. 
Ask or Assume evaluates clarification-seeking coding agents on an underspecified variant of SWE-bench Verified, studying whether agents can recognize when issue descriptions lack crucial context~\cite{edwards2026askassumeuncertaintyawareclarificationseeking}. 
These studies establish clarification as an important coding capability, but they mostly evaluate it in function-level tasks or issue-resolution settings. In such settings, clarification usually affects local behavior or a patch inside an existing repository. \rcb{} brings clarification into 0-to-1 repository generation, where missing requirements can affect architecture, public interfaces, dependencies, test protocols, and edge-case behavior. 
\section{ICAE-Bench}
\label{sec:icae_bench}

\rcb{} is an interactive benchmark for from-scratch repository generation. This section defines a benchmark instance and describes its construction. The main design principle is to expose fuzzy requirements to the agent while evaluating the final repository through reliable behavioral evidence.

\begin{figure*}[t]
\centering
\includegraphics[width=0.94\textwidth]{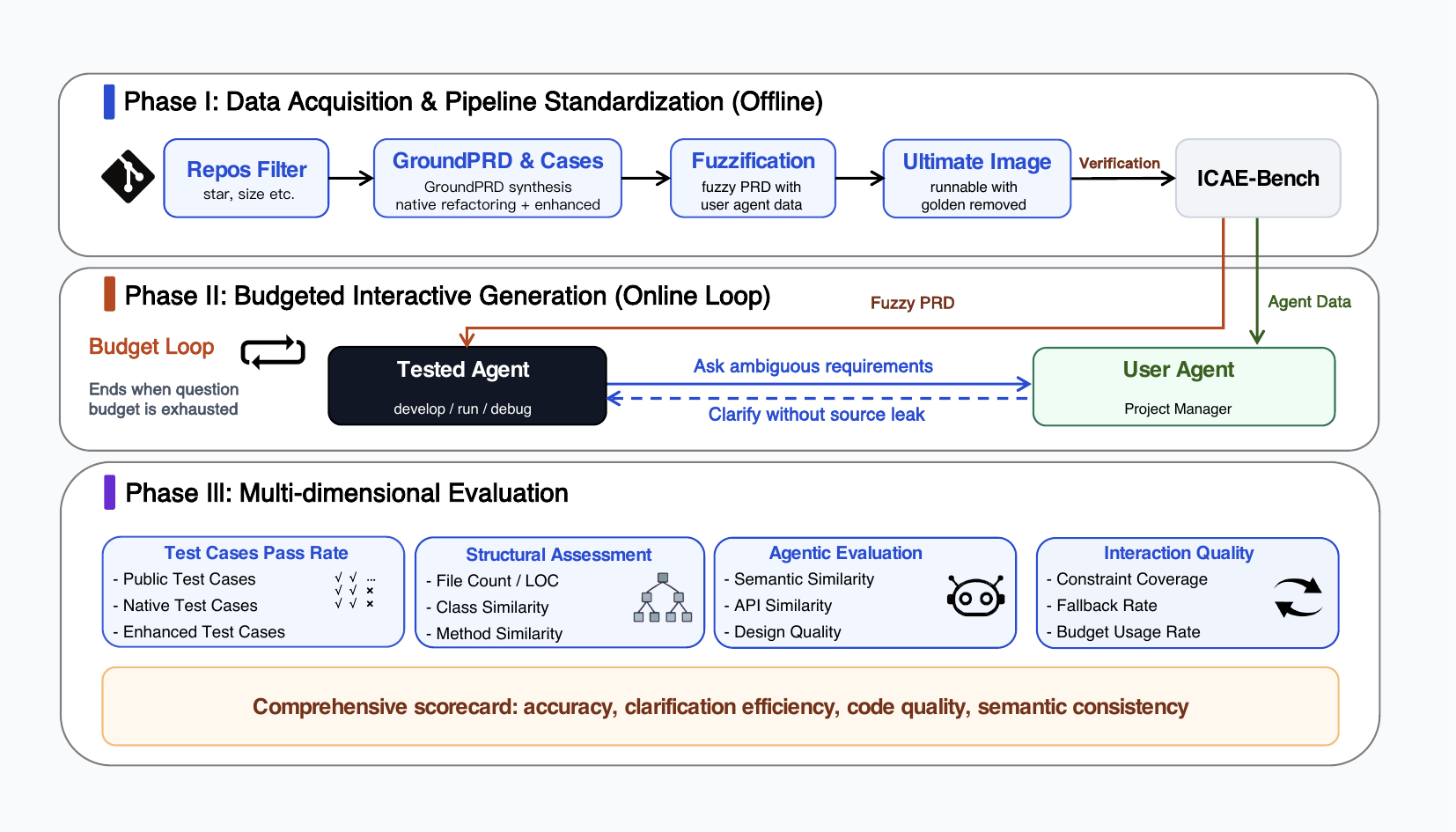}
\caption{\textbf{\rcb{} framework.} A task begins with a fuzzy PRD and an ultimate image rather than a complete implementation contract. The coding agent can develop, execute, debug, and ask the User Agent for clarifications. Final repositories are evaluated using source-based Native and Enhanced black-box cases; the Public subset is also reported as a visibility-based view.}
\label{fig:framework}
\end{figure*}

To do so, each task begins with a fuzzy PRD, allows clarification through a User Agent, and is finally scored by authoritative black-box cases that include both Public and Hidden cases. As shown in Figure~\ref{fig:framework}, there are mainly five stages: repository filtering, GroundPRD synthesis and test refactoring, PRD fuzzification, ultimate-image packaging, and artifact verification. This design gives \rcb{} realistic ambiguity, reproducible interaction, and language-agnostic evaluation. 

\subsection{Benchmark Instance Definition}
\label{sec:task-definition}

Each benchmark instance separates what the coding agent initially sees, what it may recover through interaction, and what the evaluator uses as the behavioral target. Formally, an instance is
\begin{equation}
\mathcal{T}=(D_f,E,P,U,B),
\end{equation}
where $D_f$ is the fuzzy PRD and $E$ is the ultimate-image execution environment. The Public set $P$ contains illustrative Native cases recoverable through clarification, while $B$ contains all authoritative Native and Enhanced cases. User Agent Data $U$ stores omitted constraints, the complete input/output content of $P$, and the records that ground clarification responses. The coding agent observes $(D_f,E)$, queries the User Agent backed by $U$ under a fixed budget, and produces a repository $R$ that is evaluated against $B$.

For example, in a file-parser instance, $D_f$ requests the parser without enumerating all formats or edge cases; $P$ provides representative input/output examples when requested; and $B$ additionally tests unexposed malformed and boundary inputs. Thus, requirement exposure and behavioral evaluation remain distinct.

\begin{table}[htbp]
\centering
\caption{Roles of task components in \rcb{}.}
\label{tab:component-roles}
\small
\begin{adjustbox}{width=\linewidth, center}
\begin{tabular}{ll}
\toprule
Component & Role \\
\midrule
Fuzzy PRD ($D_f$) & Initial requirement \\
Ultimate image ($E$) & Execution environment \\
Public cases ($P$) & Visible or recoverable examples \\
User Agent Data ($U$) & Interaction oracle \\
Native + Enhanced cases ($B$) & Evaluation target \\
\bottomrule
\end{tabular}
\end{adjustbox}
\end{table}

\begin{table}[t]
\centering
\caption{Overview of the \rcb{} construction and evaluation pipeline.}
\label{tab:pipeline}
\small
\begin{tabular}{p{0.24\linewidth}p{0.65\linewidth}}
\toprule
Stage & Purpose \\
\midrule
Repository filtering & Keep repositories whose original tests pass in Docker. \\
GroundPRD and case construction & Synthesize requirements and refactor tests into authoritative black-box cases. \\
PRD fuzzification & Create the fuzzy PRD and grounded User Agent Data. \\
Ultimate-image packaging & Remove solution artifacts while retaining the provisioned runtime. \\
Artifact verification & Check semantic alignment and execution consistency across all task artifacts. \\
\bottomrule
\end{tabular}
\end{table}

\subsection{Benchmark Construction}
\label{sec:construction}

We summarize the construction process in five stages. Throughout the pipeline, GroundPRD, the fuzzy PRD, User Agent Data, and all behavioral cases must describe the same underlying task. First, we retain only repositories whose original tests pass in Docker so that every source task begins from stable, reproducible behavior. Second, we refactor the original tests into standardized black-box JSON cases and synthesize GroundPRD from the verified repository behavior. Third, we fuzzify GroundPRD into an ambiguous user-facing request plus structured User Agent Data, creating a controlled gap between what is initially visible and what can be recovered through interaction. Constraint selection spans API commitments, edge-case behavior, and architectural requirements, ensuring that fuzzification does not depend on a single ambiguity type. Fourth, we package each task into an ultimate image that removes golden code and hidden artifacts while preserving the runtime environment. Fifth, we validate semantic and execution consistency across the constructed artifacts. Stage-specific prompt templates are organized in the appendices.

\paragraph{Repository Selection and Test Refactoring}
We begin with real open-source repositories that provide runnable test suites. A repository is retained only if all of its original tests pass in a controlled Docker environment, ensuring that every benchmark task originates from stable, executable behavior. For each retained repository, we record its language, popularity, file count, golden-code lines of code (LOC), and an LLM-assisted assessment of test completeness, effectiveness, reliability, and readability.

We then transform the original tests into standardized black-box cases. Each leaf feature is represented by JSON inputs and expected outputs produced by a task-specific dispatcher running against the golden implementation. Only externally observable behavioral contracts are retained; language-specific exception details are normalized into repository-level error categories. A refactored case is accepted only after it executes successfully against the golden repository in its controlled environment. These retained cases form the Native suite; the separately synthesized Enhanced suite adds robustness coverage below. Together, they decouple evaluation from the original source layout, internal APIs, and implementation language.

\paragraph{GroundPRD Synthesis and Fuzzification}
Using the verified behavior, we construct GroundPRD as a natural-language specification of the required features and general software-engineering constraints. The construction agent selects approximately one or two representative Native cases per leaf feature for readability and embeds them in GroundPRD. These examples are mirrored exactly into the Public set; the remaining retained original-test cases stay Native and hidden. Selection is judgment-based rather than a fixed count or proportion. The examples make each feature concrete and establish the test-data format without exposing source code, original tests, internal APIs, or module structure.

Fuzzification converts GroundPRD into an incomplete product request while preserving the underlying task. The levels vary initial requirement exposure while retaining a common behavioral target: L1 is the least informative, L2 restores selected L1 ambiguity points, and L3 retains most of GroundPRD while hiding a limited set of details. In Fuzzy L1, GroundPRD is rewritten into a short PM-style request and selected API, edge-case, and architectural constraints are hidden. The complete Public examples and other omitted details are transferred into User Agent Data together with trigger keywords, grounded responses, contextual pointers, and a fallback response. Fuzzy L2 restores a subset of the ambiguity points hidden in L1. Fuzzy L3, the Easy setting, uses limited deletion and light ambiguity-oriented rewriting. The hierarchy is operational rather than a claim that ambiguity has a single universal scale; it controls how much task information is available before interaction. Because ambiguity points and their retrieval records are generated together, omitted information remains recoverable through sufficiently targeted questions.

\begin{figure}[t]
\centering
\begin{tikzpicture}[
  node distance=1.2mm,
  level/.style={draw, rounded corners=1.2pt, align=center, text width=0.82\linewidth, minimum height=5.2mm, font=\footnotesize},
  flow/.style={-{Latex[length=1.3mm]}, line width=0.5pt}
]
\node[level] (ground) {GroundPRD: complete specification};
\node[level, below=of ground] (l3) {Fuzzy L3: limited deletion and light rewriting};
\node[level, below=of l3] (l2) {Fuzzy L2: selected L1 ambiguity points restored};
\node[level, below=of l2] (l1) {Fuzzy L1: least initial requirement information};
\draw[flow] (l1) -- (l2);
\draw[flow] (l2) -- (l3);
\draw[flow] (l3) -- (ground);
\end{tikzpicture}
\caption{Initial-requirement hierarchy. Information available before interaction increases from Fuzzy L1 to GroundPRD, while the behavioral target remains fixed.}
\label{fig:prd-hierarchy}
\end{figure}
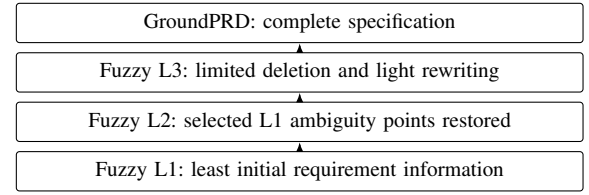

\paragraph{Enhanced Behavioral Cases}
Native cases preserve behavior covered by the original repository tests, but those tests may underrepresent difficult boundaries. We therefore synthesize Enhanced cases targeting boundary values, special constraints, stress behavior, and logic-specific corner cases. Enhanced cases must be disjoint from Native cases, and every synthesized case is executed against the golden implementation before acceptance. This adds robustness coverage without requiring generated repositories to reproduce the original implementation structure.

\paragraph{Ultimate Image Generation}
\label{sec:ultimate-image-generation}
The \emph{base image} is the official language-specific image downloaded from Docker Hub. We favor mainstream stable or LTS versions to maximize compatibility with real repositories while keeping the environment specification explicit and reproducible (Table~\ref{tab:image-versions}). To construct the \emph{ultimate image}, we run the golden repository successfully in the base image, install the dependencies needed by that repository, and then remove the golden code, original tests, and hidden construction artifacts. The result is a fully provisioned execution environment without solution artifacts: it retains the verified dependency stack while forcing the agent to rebuild the repository and its runnable interface from the exposed requirement information alone. For Kotlin, we use a Kotlin 1.9.25 environment built on Eclipse Temurin 17 and describe it generically to avoid deanonymizing any private image registry.

\begin{table}[htbp]
\centering
\caption{Language-specific execution environment images used for benchmark packaging.}
\label{tab:image-versions}
\small
\begin{adjustbox}{width=\linewidth, center}
\begin{tabular}{lll}
\toprule
Language & Runtime Version & Base Image \\
\midrule
C\# & .NET 8 & \texttt{mcr.microsoft.com/dotnet/sdk:8.0} \\
C++ & GCC 12 & \texttt{gcc:12} \\
Dart & Dart 3.5 & \texttt{dart:3.5} \\
Go & Go 1.22 & \texttt{golang:1.22} \\
Java & Java 17 & \texttt{eclipse-temurin:17} \\
JavaScript & Node 20 & \texttt{node:20} \\
TypeScript & Node 20 & \texttt{node:20} \\
Kotlin & Kotlin 1.9.25 & \texttt{Eclipse Temurin 17-based custom image} \\
PHP & PHP 8.2 & \texttt{php:8.2} \\
Python & Python 3.11 & \texttt{python:3.11} \\
Ruby & Ruby 3.2 & \texttt{ruby:3.2} \\
Rust & Rust 1.81 & \texttt{rust:1.81} \\
\bottomrule
\end{tabular}
\end{adjustbox}
\end{table}

\subsection{Artifact Verification}
\label{sec:artifact-verification}

Verification prevents benchmark-construction errors from being misattributed to coding agents. Repository-level executability is enforced upstream: retained golden repositories must pass their original and reconstructed Native cases, and every Enhanced case must pass against the golden implementation before acceptance.

We then establish GroundPRD correctness through iterative reconstruction in the ultimate image. Claude-Opus-4.8 implements each repository from GroundPRD, and we inspect failing Public or Native cases. We repair incorrect cases or revise an omitted or misstated contract, repeating the process until GroundPRD, Public examples, and Native cases are mutually consistent.

Finally, we verify that the fuzzy PRD and User Agent Data preserve the validated GroundPRD contract using two complementary checks. First, GPT-5.5 jointly compares the fuzzy PRD and User Agent Data with GroundPRD under a fixed rubric for semantic similarity, equivalence, and execution-contract risk. On \rcblite{}, Fuzzy L1 and L3 are equivalent to GroundPRD for all 50 repositories, with mean similarity scores of 0.952 and 0.942 and no case judged likely to fail the same tests. Second, black-box execution-equivalence checks validate the contract against the golden implementation. The two checks test semantic alignment and executable behavior, respectively. Prompts are provided in Appendices~\ref{appendix:fuzzy-prd-semantic} and~\ref{appendix:fuzzy-prd-easy-prompts}.

\paragraph{Generation Harness}
At evaluation time, the harness provisions a fresh container from the task image, exposes only the task-facing starter materials, and asks the coding agent to implement the repository entirely inside that container. A common interface integrates Claude Code and OpenHands while preserving their native editing and execution workflows. The prompt requires the agent to build its own execution adapter, use any Public examples recovered through interaction for self-checking, and make the repository reproducibly runnable. Final scoring replays the generated repository in a fresh container from the same image, so transient development-time actions count only when reflected in the delivered artifacts. The full instruction is included in Appendix~\ref{appendix:agent-task}.

\begin{table}[htbp]
\centering
\caption{Dataset statistics. Token lengths are computed over task documents and User Agent Data.}
\label{tab:dataset-statistics}
\begin{tabular}{lcc}
\toprule
\multirow{2}{*}{\textbf{Statistic}} & \multicolumn{2}{c}{\textbf{Count}} \\
\cmidrule(lr){2-3}
 & \textbf{\rcb{}} & \textbf{\rcblite{}} \\
\midrule
\textbf{GroundPRD Tokens} \\
~~~~- \textit{Maximum} & $43,990$ & $43,990$ \\
~~~~- \textit{Minimum} & $1,767$ & $1,767$ \\
~~~~- \textit{Average} & $6,764$ & $6,170$ \\
\midrule
\textbf{Fuzzy PRD Tokens} \\
~~~~- \textit{Maximum} & $367$ & $327$ \\
~~~~- \textit{Minimum} & $181$ & $226$ \\
~~~~- \textit{Average} & $276$ & $273$ \\
\midrule
\textbf{User Agent Data Tokens} \\
~~~~- \textit{Maximum} & $53,887$ & $53,887$ \\
~~~~- \textit{Minimum} & $515$ & $648$ \\
~~~~- \textit{Average} & $7,091$ & $6,945$ \\
\midrule
\textbf{Languages} & $12$ & $10$ \\
\midrule
\textbf{Total Tasks} & $480$ & $50$ \\
\bottomrule
\end{tabular}
\end{table}

\subsection{Benchmark Statistics}
\label{sec:statistics}

\begin{figure}[t]
\centering
\includegraphics[width=\linewidth]{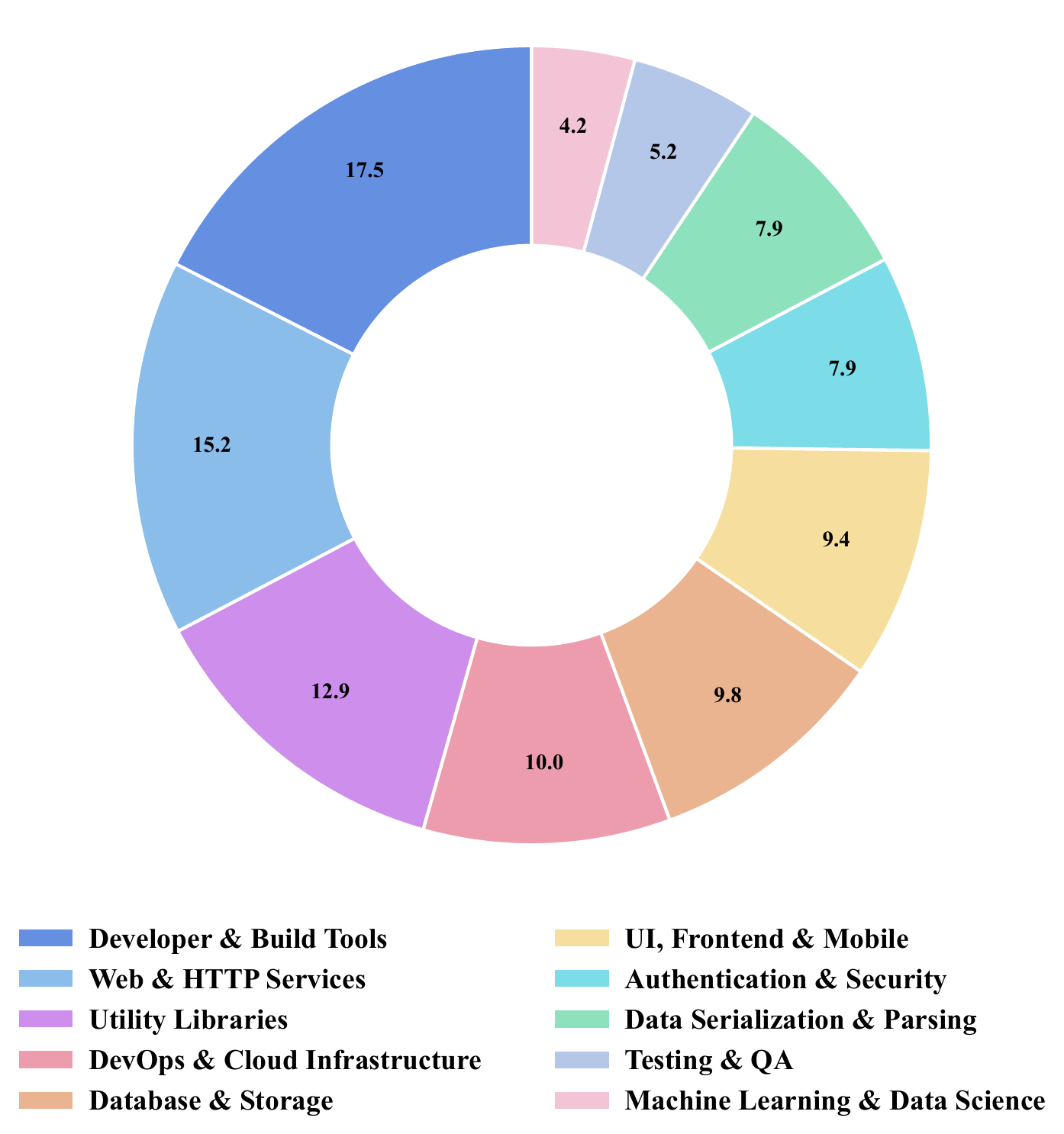}
\caption{Task category distribution in \rcb{} (total=480).}
\label{fig:categories}
\end{figure}

\begin{figure}[htbp]
\centering
\includegraphics[width=0.7\linewidth]{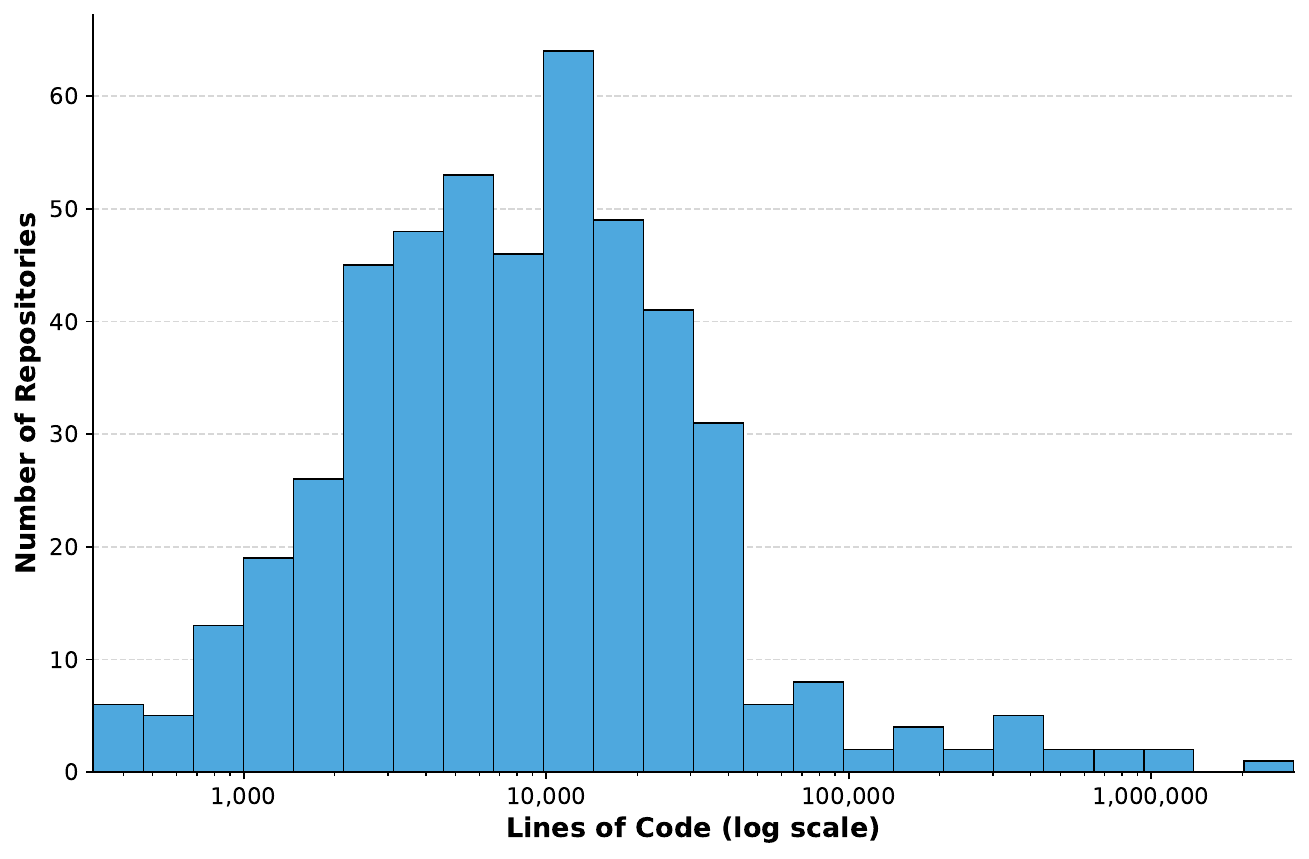}
\caption{Task golden-code LOC distribution in \rcb{} (total=480).}
\label{fig:repo_loc}
\end{figure}

As shown in Table~\ref{tab:dataset-statistics} and Figure~\ref{fig:categories}, \rcb{} covers 12 programming languages: C\#, C++, Dart, Go, Java, JavaScript, Kotlin, PHP, Python, Ruby, Rust, and TypeScript. The full benchmark contains 40 tasks per language. \rcblite{} is selected from \rcb{} by prioritizing smaller repositories according to golden-code LOC, making it suitable for rapid ablations and framework comparisons. Figure~\ref{fig:repo_loc} shows the full benchmark's golden-code LOC distribution.

\paragraph{Golden LOC.}
Maximum: $2,918,810$ / $1,509$; minimum: $318$ / $318$; average: $35,533$ / $979$ for \rcb{} / \rcblite{}, respectively.

\paragraph{Golden File Count.}
Maximum: $17,488$ / $61$; minimum: $4$ / $4$; average: $258$ / $17$ for \rcb{} / \rcblite{}, respectively.

\subsection{Interactive User Agent}
\label{sec:user-agent}

The User Agent is a model-based router that simulates a strict product stakeholder. Our default configuration uses DeepSeek-V3.2 and a budget of 16 queries. The coding agent sends a task identifier and one natural-language question to an HTTP endpoint; the server loads that task's User Agent Data, invokes the router, and returns a natural-language reply together with the remaining budget. Each reply addresses at most three matched technical points, after which the coding agent may issue the next query. The full request format and initialization prompt are provided in Appendices~\ref{appendix:suffix} and~\ref{appendix:user-agent}.

Routing is restricted to benchmark-authored records. During fuzzification, each omitted ambiguity point is assigned an identifier, semantically equivalent trigger phrases, a grounded response, and optional context pointers; omitted Public examples and a fixed fallback response are stored in the same task data. For each query, the router may return only responses associated with matched records or the fallback. Its structured internal log records matched constraint identifiers, interface-alignment events, and fallback use.

This protocol limits leakage at three boundaries. The User Agent receives no golden source code, original tests, repository identity, or hidden evaluation cases; its task knowledge is limited to the generated records. The coding agent receives only the router's natural-language reply, not the records or internal log. Each tested agent develops in an isolated Docker container without authoritative cases; those cases are injected only into a fresh evaluation container after generation. Thus, interaction can reveal omitted requirements and Public examples, but not solution artifacts or hidden test contents.

\subsection{Evaluation Metrics}
\label{sec:metrics}

\paragraph{Functional Correctness}
Functional correctness is the primary measure of whether a generated repository implements the required externally observable behavior. We organize cases along two orthogonal axes. By source, \emph{Native} cases are transformed from original repository tests, whereas \emph{Enhanced} cases add synthesized robustness coverage. By visibility, \emph{Public} cases are exposed in GroundPRD and \emph{Hidden} cases are all remaining cases. Public is a subset of Native, so $\mathrm{Hidden}=(\mathrm{Native}\cup\mathrm{Enhanced})\setminus\mathrm{Public}$; Public/Hidden are visibility partitions, whereas Native/Enhanced are source partitions. Fuzzy L1/L2 omit Public cases from the initial PRD, but they remain recoverable through interaction. Tables~\ref{tab:main-results-full} and~\ref{tab:main-results-lite} report Public, Native, and Enhanced views. \emph{Overall} is the case-level pass rate over all Native and Enhanced cases, computed as $(\#\text{passed Native}+\#\text{passed Enhanced})/(\#\text{Native}+\#\text{Enhanced})$. Evaluation is host-verified: a fresh container receives the authoritative cases, runs the agent-produced entry point, and returns outputs for host-side comparison.

\paragraph{Agentic Evaluation}
Functional tests cannot distinguish a narrowly hard-coded solution from a maintainable repository. We therefore use static critic-model review over the generated and golden source trees. Semantic similarity measures alignment of behavior and abstractions; API similarity compares importable types, functions, names, signatures, and coverage; and design quality assesses the generated repository's architecture, including separation of core logic from the execution adapter. These independent scores diagnose behavioral breadth, interface fidelity, and engineering quality, respectively. The critic prompt is included in Appendix~\ref{appendix:critic}.

\paragraph{Structural Assessment}
Repository-scale failures may be hidden by case-level pass rates. We therefore compare generated and golden repositories using file-count and LOC ratios plus class, method, and namespace/package similarities. These diagnostics reveal under-building, over-building, or collapsing a multi-module project into a monolithic adapter; they do not reward literal reproduction of the golden implementation.

\paragraph{Interaction Quality}
Interaction metrics measure whether an agent retrieves relevant hidden requirements efficiently. From the router logs, let $C$ be the hidden constraint identifiers and $H$ the unique identifiers matched at least once. Constraint coverage is $|H|/|C|$; fallback rate is the fraction of queries matching no grounded record; and budget usage is the issued-query count divided by the allowed budget. We macro-average task-level rates so tasks with more constraints do not dominate.

Pass rates, agentic scores, structural similarities, and constraint coverage are higher-is-better within their respective dimensions; fallback rate is lower-is-better, and budget usage is descriptive. File-count and LOC ratios may exceed 100\%; values near 100\% indicate similar scale, not necessarily better quality.

\section{Experiments}
\label{sec:experiments}

We design experiments to answer four research questions. \textbf{RQ1:} How well can current coding agents implement repositories from fuzzy requirements (Section~\ref{sec:main-results})? \textbf{RQ2:} How much of the gap to GroundPRD can interaction recover (Section~\ref{sec:main-results})? \textbf{RQ3:} Where do bottlenecks arise in the clarification-to-implementation pipeline (Section~\ref{sec:analysis})? \textbf{RQ4:} How sensitive are results to the agent framework, User Agent backbone, interaction budget, test scaffolding, thinking configuration, execution environment, and implementation language (Section~\ref{sec:analysis})?

\subsection{Experimental Settings}
\label{sec:experimental-settings}

\paragraph{Datasets and metrics.}
We report the main comparison on the full \rcb{} benchmark and use \rcblite{}, a 50-repository subset selected by sorting the full tasks by golden-code LOC, for repeated ablations. The full split measures realistic repository-scale performance, while \rcblite{} enables controlled comparisons across model backbones, agent frameworks, User Agent backbones, interaction budgets, thinking-budget settings, and execution environments.

We use the functional, agentic, structural, and interaction metrics defined in Section~\ref{sec:metrics}; all reported values are percentages unless noted otherwise.

\paragraph{Models and execution.}
The main \rcblite{} and full \rcb{} comparisons evaluate six coding models under the Claude Code framework~\cite{anthropic_claude_agent_sdk_2026}: GPT-5.5~\cite{singh2025openai}, Claude-Opus-4.8~\cite{anthropic_claude_opus_4_8_2026}, Claude-Sonnet-4.6~\cite{anthropic_claude_sonnet_4_6_2026}, GLM-5.1~\cite{zeng2026glm}, Gemini-3.1-Pro~\cite{google_deepmind_gemini_3_1_pro_2026}, and MiniMax-M2.5~\cite{chen2026minimax}. Unless otherwise specified, the main result tables report the adaptive thinking setting, where the model decides its own thinking length. For GLM-5.1, we additionally analyze a think-8k setting that caps the maximum thinking length at 8k tokens. The tested agent receives a fuzzy PRD, may ask at most 16 questions, and interacts with a DeepSeek-V3.2 User Agent that routes questions to grounded hidden constraints rather than generating free-form answers. GroundPRD runs remove ambiguity by providing the complete requirement document upfront and therefore serve as an upper-bound reference. Although tasks are derived from real open-source repositories, the evaluated agents do not see original repository identities, original source code, original tests, or hidden implementation details; they see only benchmark-authored task materials and any clarification returned through grounded User Agent records.

\subsection{Main Results}
\label{sec:main-results}

The main results address two questions in sequence. To answer RQ1, we first evaluate the default Fuzzy L1 setting with User Agent interaction on the full benchmark. This baseline alone cannot separate missing requirements from weak implementation. To answer RQ2, we vary \emph{requirement exposure}---the amount and organization of requirement information available before coding---over the same tasks, from Fuzzy L1--L3 and RecoveredPRD to the complete GroundPRD.

Under the shared default framework and interaction protocol, Table~\ref{tab:main-results-full} compares six coding models on the full benchmark, while Table~\ref{tab:main-results-lite} reports the corresponding \rcblite{} results.

\begin{table*}[t]
\centering
\caption{Main results on the full \rcb{} split using Claude Code. Values are percentages; File and LOC are generated-to-golden ratios and may exceed 100.}
\label{tab:main-results-full}
\small
\begin{tabular}{lccccccc}
\toprule
\multirow{2}{*}{Model} & \multicolumn{4}{c}{Test Cases Pass Rate} & \multicolumn{3}{c}{Agentic Evaluation} \\
\cmidrule(lr){2-5}\cmidrule(lr){6-8}
 & Overall & Public & Native & Enhanced & Sem. & API & Design \\
\midrule
Claude-Opus-4.8 & \textbf{38.2} & 48.5 & 41.4 & \textbf{35.5} & 22.6 & \textbf{12.1} & \textbf{44.4} \\
GPT-5.5 & 37.2 & \textbf{50.3} & \textbf{42.0} & 32.8 & 21.5 & 9.3 & 36.5 \\
Gemini-3.1-Pro & 27.0 & 37.0 & 30.7 & 23.5 & 18.9 & 8.9 & 28.6 \\
GLM-5.1 & 26.6 & 36.8 & 30.0 & 23.7 & 21.1 & 10.0 & 37.1 \\
Claude-Sonnet-4.6 & 21.8 & 29.0 & 24.1 & 19.4 & \textbf{22.9} & 10.1 & 37.5 \\
MiniMax-M2.5 & 0.8 & 1.5 & 1.2 & 0.6 & 11.8 & 5.5 & 23.1 \\
\midrule
\multirow{2}{*}{Model} & \multicolumn{4}{c}{Structural Assessment} & \multicolumn{3}{c}{Interaction Quality} \\
\cmidrule(lr){2-5}\cmidrule(lr){6-8}
 & File & LOC & Class & Method & Constr. & Fallback & Budget \\
\midrule
Claude-Opus-4.8 & 337.6 & 981.1 & \textbf{17.5} & \textbf{10.8} & 69.6 & 21.6 & 81.5 \\
GPT-5.5 & 44.0 & 154.9 & 14.6 & 9.0 & \textbf{73.7} & \textbf{20.4} & 76.8 \\
Gemini-3.1-Pro & 46.2 & 221.9 & 15.9 & 10.4 & 56.3 & 28.1 & 74.2 \\
GLM-5.1 & 12.1 & 15.9 & 15.0 & 9.6 & 63.9 & 28.1 & 95.0 \\
Claude-Sonnet-4.6 & 15.5 & 15.9 & 17.0 & 10.4 & 57.7 & 24.2 & 76.0 \\
MiniMax-M2.5 & 41.2 & 117.0 & 13.1 & 7.5 & 44.3 & 28.5 & 70.6 \\
\bottomrule
\end{tabular}
\end{table*}

\begin{table*}[htbp]
\centering
\caption{Main results on \rcblite{} using Claude Code. Values are percentages; File and LOC are generated-to-golden ratios and may exceed 100.}
\label{tab:main-results-lite}
\small
\begin{tabular}{lccccccc}
\toprule
\multirow{2}{*}{Model} & \multicolumn{4}{c}{Test Cases Pass Rate} & \multicolumn{3}{c}{Agentic Evaluation} \\
\cmidrule(lr){2-5}\cmidrule(lr){6-8}
 & Overall & Public & Native & Enhanced & Sem. & API & Design \\
\midrule
Claude-Opus-4.8 & 48.2 & 56.6 & 52.9 & 44.4 & 28.5 & 14.3 & \textbf{49.1} \\
GPT-5.5 & \textbf{53.3} & \textbf{63.8} & \textbf{60.9} & \textbf{47.4} & 30.0 & 15.2 & 48.7 \\
Gemini-3.1-Pro & 36.6 & 45.5 & 41.4 & 33.1 & 27.2 & 14.2 & 31.4 \\
GLM-5.1 & 40.0 & 47.6 & 44.5 & 35.9 & 22.6 & 11.0 & 37.9 \\
Claude-Sonnet-4.6 & 40.2 & 47.0 & 44.1 & 37.1 & \textbf{34.2} & \textbf{16.1} & 44.3 \\
MiniMax-M2.5 & 2.8 & 2.7 & 4.4 & 1.9 & 19.2 & 8.2 & 28.6 \\
\midrule
\multirow{2}{*}{Model} & \multicolumn{4}{c}{Structural Assessment} & \multicolumn{3}{c}{Interaction Quality} \\
\cmidrule(lr){2-5}\cmidrule(lr){6-8}
 & File & LOC & Class & Method & Constr. & Fallback & Budget \\
\midrule
Claude-Opus-4.8 & 45.3 & 47.1 & \textbf{36.8} & 17.7 & 67.8 & \textbf{17.9} & 75.5 \\
GPT-5.5 & 36.8 & 35.0 & 32.6 & 17.7 & \textbf{73.2} & 21.0 & 71.7 \\
Gemini-3.1-Pro & 62.2 & 743.3 & 35.4 & 18.0 & 58.0 & 23.5 & 68.5 \\
GLM-5.1 & 32.4 & 64.5 & 33.2 & 17.4 & 63.3 & 28.4 & 95.4 \\
Claude-Sonnet-4.6 & 33.9 & 51.0 & 34.9 & \textbf{18.4} & 60.8 & 25.9 & 75.1 \\
MiniMax-M2.5 & 17.9 & 32.5 & 30.8 & 12.6 & 48.1 & 28.1 & 69.1 \\
\bottomrule
\end{tabular}
\end{table*}

Table~\ref{tab:language-results-full} reports the balanced programming-language breakdown. The four metric families expose distinct capabilities. Functionally, Public rates consistently exceed Enhanced rates, indicating difficulty beyond visible examples. Agentic scores do not mirror the functional ranking: Claude-Sonnet-4.6 has the highest semantic score but ranks fifth in Overall. Structural ratios reveal sharply different construction strategies; Claude-Opus-4.8 produces much larger repositories than GPT-5.5, despite their similar Overall rates. Interaction metrics show a further separation: GPT-5.5 recovers more constraints than Claude-Opus-4.8 yet scores one point lower Overall. Neither repository scale nor recovered information is therefore a proxy for correctness.

\begin{table}[t]
\centering
\caption{Per-language overall pass rate on the full \rcb{} split.}
\label{tab:language-results-full}
\small
\begin{adjustbox}{width=\linewidth, center}
\begin{tabular}{lcccccc}
\toprule
Model & C\# & C++ & Dart & Go & Java & JavaScript \\
\midrule
Claude-Opus-4.8 & 39.9 & 29.5 & 38.7 & 44.8 & 39.1 & 34.4 \\
GPT-5.5 & 39.4 & 31.5 & 40.6 & 39.7 & 38.4 & 34.9 \\
Gemini-3.1-Pro & 18.6 & 13.8 & 26.7 & 22.0 & 35.0 & 33.2 \\
GLM-5.1 & 23.2 & 26.3 & 33.4 & 21.9 & 20.8 & 27.6 \\
Claude-Sonnet-4.6 & 23.6 & 19.1 & 21.9 & 30.1 & 19.3 & 18.9 \\
MiniMax-M2.5 & 0.0 & 1.0 & 0.9 & 0.1 & 0.0 & 0.2 \\
\midrule
Model & Kotlin & PHP & Python & Ruby & Rust & TypeScript \\
\midrule
Claude-Opus-4.8 & 39.3 & 36.5 & 33.0 & 38.9 & 40.3 & 43.7 \\
GPT-5.5 & 31.7 & 42.2 & 34.9 & 41.4 & 37.4 & 34.6 \\
Gemini-3.1-Pro & 26.8 & 35.9 & 31.4 & 24.9 & 22.7 & 32.7 \\
GLM-5.1 & 29.9 & 38.4 & 26.8 & 19.4 & 23.9 & 27.3 \\
Claude-Sonnet-4.6 & 22.7 & 24.1 & 23.6 & 15.7 & 21.4 & 20.5 \\
MiniMax-M2.5 & 2.1 & 0.2 & 4.5 & 0.3 & 0.0 & 0.9 \\
\bottomrule
\end{tabular}
\end{adjustbox}
\end{table}

\textbf{Answer to RQ1.} Current agents implement a nontrivial fraction of fuzzy requirements, but the best Overall rate is only 38.2\%. The leading model varies across programming languages, and similar functional scores conceal substantial differences in artifact design, repository scale, and requirement recovery. The consistent Public--Enhanced gap further shows that reproducing visible examples does not ensure robust behavior.

\begin{figure}[t]
\centering
\includegraphics[width=\linewidth]{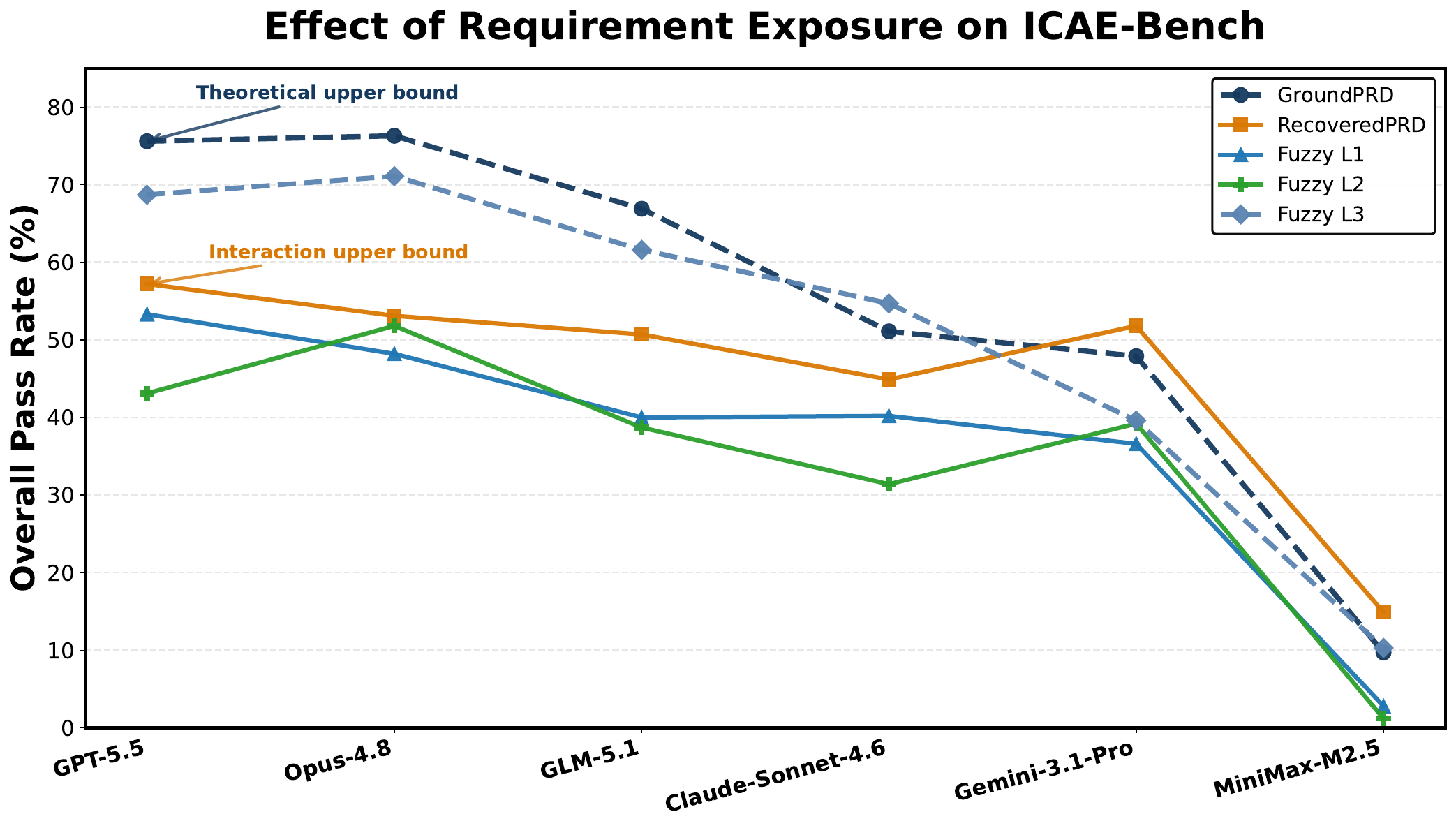}
\caption{Overall pass rate across PRD ambiguity settings on \rcblite{}. Fuzzy L2 restores a subset of the ambiguity points hidden in L1; Fuzzy L3 preserves more of GroundPRD through limited deletion and light ambiguity-oriented rewriting.}
\label{fig:prd-levels-plot}
\end{figure}

\begin{figure}[htbp]
\centering
\includegraphics[width=\linewidth]{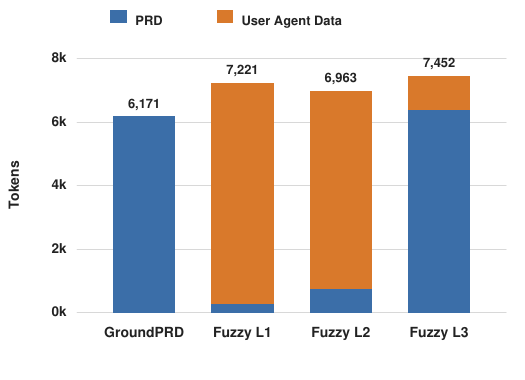}
\caption{Requirement-token comparison on \rcblite{}. User Agent Data is accessible only through interaction.}
\label{fig:prd-token-comparison}
\end{figure}

Figure~\ref{fig:prd-levels-plot} compares three fuzzy levels, RecoveredPRD, and GroundPRD over the same tasks. RecoveredPRD is constructed by appending all L1 question--answer records to the Fuzzy L1 PRD before generation, thereby removing retrieval while preserving their fragmented form. L2 restores selected L1 ambiguity points, L3 retains most of GroundPRD, and GroundPRD provides the complete coherent specification. Their semantic alignment is validated in Section~\ref{sec:artifact-verification}; Figure~\ref{fig:prd-token-comparison} quantifies L1--L3 exposure.

\begin{figure}[t]
\centering
\includegraphics[width=\linewidth]{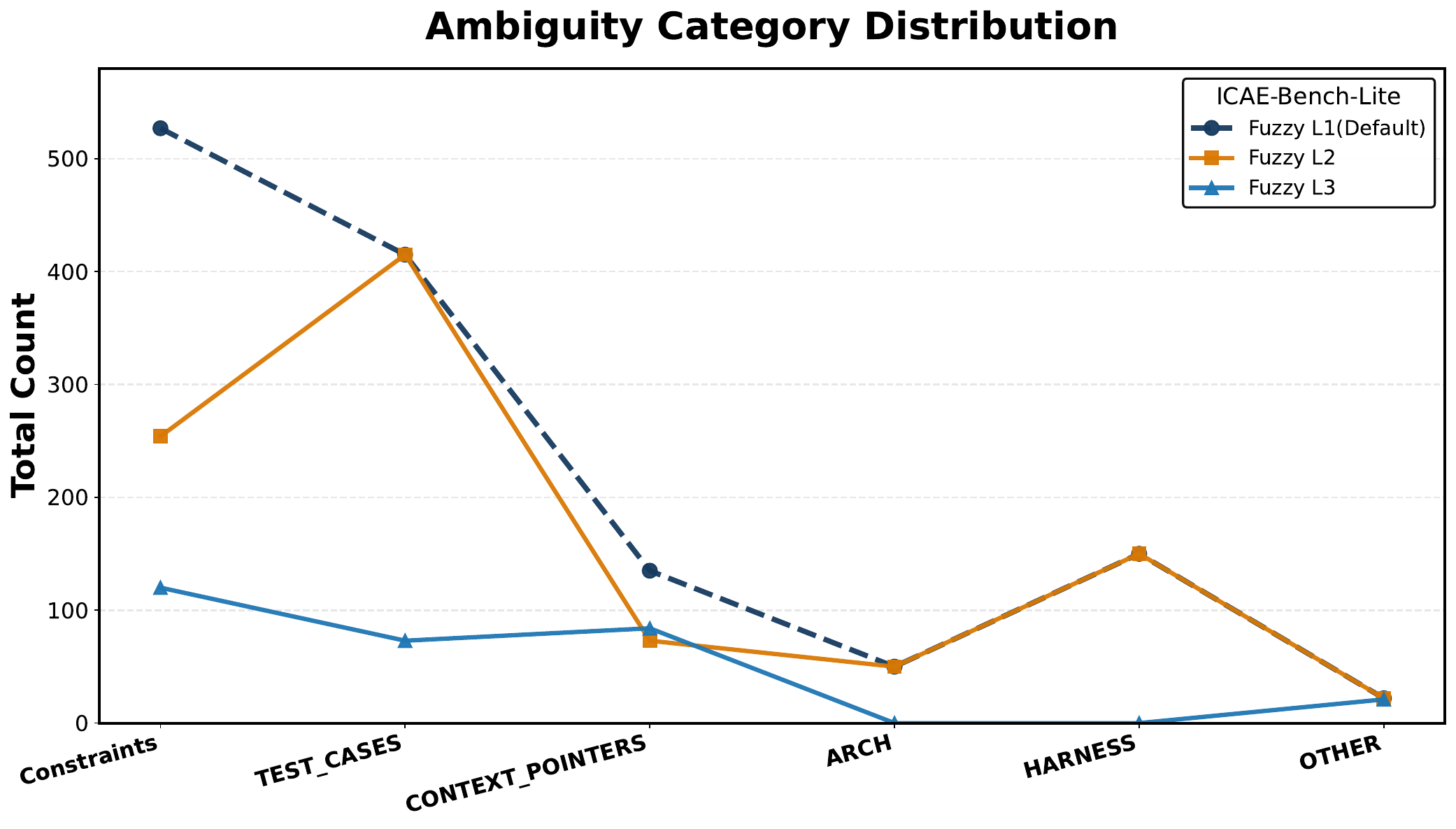}
\caption{Distribution of ambiguity categories across Fuzzy L1--L3 in \rcb{}.}
\label{fig:ambiguity-category-distribution}
\end{figure}

GroundPRD performs best for four of six models. RecoveredPRD remains below it for GPT-5.5, Claude-Opus-4.8, GLM-5.1, and Claude-Sonnet-4.6, despite exposing every recoverable L1 record. Gemini-3.1-Pro and MiniMax-M2.5 are exceptions, and the fuzzy-level trends are not uniformly monotonic. Thus, retrieval explains only part of the gap: agents must also organize fragmented answers, retain them, and convert them into implementation decisions. Figure~\ref{fig:ambiguity-category-distribution} confirms that this challenge spans multiple ambiguity categories rather than one missing-information type.

\textbf{Answer to RQ2.} Interaction and greater initial exposure recover only part of the gap to GroundPRD, with substantial variation across models. Even complete access to recoverable records is insufficient when those records are not organized and used effectively during implementation.

Together, RQ1 and RQ2 expose two distinct gaps. The \emph{requirement-access gap} separates a fuzzy PRD from the information recoverable through clarification; the \emph{information-to-execution gap} separates recovered requirements from a correct repository. The first motivates interaction, whereas the second concerns retention, implementation, and verification over a long development trajectory. The following analyses focus on where this second gap emerges and which components can reduce it.

\subsection{Analysis}
\label{sec:analysis}

The main results show that additional requirement information recovers only part of the performance gap. To answer RQ3, evaluator-visible failure categories localize breakdowns, while a controlled executable-feedback intervention tests whether verification reduces them. We then answer RQ4 by varying the interaction stack, reasoning configuration, and execution setting.

\subsubsection{Failure Mode Taxonomy}
\label{sec:failure-modes}

\begin{figure}[htbp]
\centering
\includegraphics[width=\linewidth]{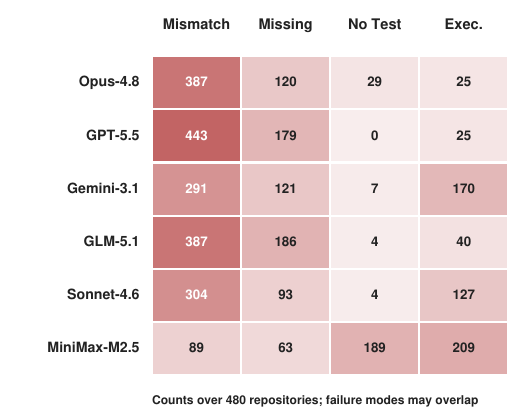}
\caption{Evaluator-visible outcomes on full \rcb{} main runs. Failure modes are not mutually exclusive.}
\label{fig:failure-modes}
\end{figure}

Pass rate alone cannot locate a breakdown. Figure~\ref{fig:failure-modes} therefore reports four evaluator-visible failure modes. \emph{Mismatch} means that a case produced output that differed from the expected result, indicating a logic or behavioral-correctness failure. \emph{Missing} means that some cases produced no output while the suite otherwise ran, indicating incomplete feature coverage. \emph{Exec.} means that the harness ran but all outputs were absent because of a recognized build or runtime error, indicating buildability, dependency, or environment failure. \emph{No Test} means that the agent did not produce the required \texttt{rcb\_tests/test.sh} entry point, indicating instruction-following or harness-compliance failure.

Failure labels are assigned per Public, Native, and Enhanced suite. A repository may therefore exhibit different modes across suites, and a single suite may contain both missing and mismatched cases; the columns are not mutually exclusive. A \emph{Clean} repository passes every evaluated case. We omit this success category from the failure-mode figure; only 1--13 of 480 repositories per model are Clean because this repository-level criterion is stricter than the case-level Overall pass rate. Constraint forgetting and interaction inefficiency are not directly identifiable from execution outcomes and are analyzed through interaction metrics and controlled ablations.

The counts expose distinct bottlenecks. Stronger models more often reach execution but violate the behavioral contract, whereas weaker models more often omit the harness or fail during execution. Missing outputs further isolate incomplete feature coverage from incorrect implemented behavior.

\subsubsection{Effect of Test Scaffolding}
\label{sec:test-scaffolding}

Failure categories do not reveal whether clarified information is difficult to apply and verify. We therefore provide ready-to-run Public case files while keeping the same case content recoverable through the User Agent in both settings. Overall rises from 37.4\% to 61.8\%, while constraint coverage changes from 63.8\% to 61.2\% (Table~\ref{tab:test-scaffold}).

\begin{table}[t]
\centering
\caption{Effect of placing Public case files in the workspace for GLM-5.1 think-8k on \rcblite{}. \textbf{w/} includes the files; \textbf{w/o} (default) omits them while retaining interaction-based access to the same case content.}
\label{tab:test-scaffold}
\small
\begin{tabular}{lcccc}
\toprule
Setting & Pass OA & Design & Constr. & Fallback \\
\midrule
w/ & \textbf{61.8} & 38.5 & 61.2 & 29.7 \\
w/o (default) & 37.4 & \textbf{43.2} & \textbf{63.8} & \textbf{27.7} \\
\bottomrule
\end{tabular}
\end{table}

The treatment changes the representation and actionability of information, not its semantic coverage. Public files package inputs, expected outputs, and the invocation format as an executable structure; without them, the agent must recover fragments through questions and reconstruct the same verification scaffold. Immediate execution and debugging feedback therefore explain the gain more directly than additional requirements. The improvement also extends to Native and Enhanced behavior despite slightly lower constraint coverage.

\textbf{Answer to RQ3.} The analyses in Sections~\ref{sec:failure-modes} and~\ref{sec:test-scaffolding} locate breakdowns at both execution and behavioral fidelity and show that verification can improve correctness without recovering more constraints. Low scores therefore reflect failures across the clarification-to-implementation chain rather than requirement access alone.


\subsubsection{Impact of Agent Frameworks}
\label{sec:framework}

RQ3 shows that obtaining requirements is not enough; the surrounding framework may determine whether they survive implementation. Because frameworks control editing, execution, and state tracking, Table~\ref{tab:framework} tests the same models and tasks under Claude Code and OpenHands~\cite{wang2025openhands}. OpenHands lowers Overall for every model by 5.5--21.8 points, although GPT-5.5 recovers more constraints.

\begin{table}[t]
\centering
\caption{Agent-framework comparison on \rcblite{}.}
\label{tab:framework}
\small
\begin{adjustbox}{width=\linewidth, center}
\begin{tabular}{lcccc}
\toprule
Model & Pass OA & Design & Constr. & Fallback \\
\midrule
\multicolumn{5}{c}{\textbf{\textit{Claude Code}}} \\
\midrule
GPT-5.5 & \textbf{53.3} & 48.7 & 73.2 & 21.0 \\
Claude-Opus-4.8 & 48.2 & 49.1 & 67.8 & 17.9 \\
Claude-Sonnet-4.6 & 40.2 & 44.3 & 60.8 & 25.9 \\
GLM-5.1 & 37.4 & 43.2 & 63.8 & 27.7 \\
Gemini-3.1-Pro & 36.6 & 31.4 & 58.0 & 23.5 \\
MiniMax-M2.5 & 2.8 & 28.6 & 48.1 & 28.1 \\
\midrule
\multicolumn{5}{c}{\textbf{\textit{OpenHands}}} \\
\midrule
GPT-5.5 & 31.5 & 39.4 & \textbf{76.8} & 20.0 \\
Claude-Opus-4.8 & 42.7 & \textbf{56.6} & 68.6 & \textbf{13.1} \\
Claude-Sonnet-4.6 & 23.6 & 44.3 & 55.5 & 15.2 \\
GLM-5.1 & 28.4 & 43.1 & 72.0 & 24.3 \\
Gemini-3.1-Pro & 28.6 & 34.8 & 62.2 & 25.9 \\
MiniMax-M2.5 & 1.1 & 31.7 & 43.0 & 27.6 \\
\bottomrule
\end{tabular}
\end{adjustbox}
\end{table}

Framework sensitivity is model dependent. GPT-5.5, the strongest model under Claude Code, drops by 21.8 points and falls behind Claude-Opus-4.8 under OpenHands; Claude-Opus-4.8 drops by only 5.5 points. Thus, framework choice changes both absolute performance and ranking. GPT-5.5's higher OpenHands constraint coverage does not offset weaker downstream execution, underscoring the role of reliable editing, command execution, state tracking, and termination.

\subsubsection{Impact of User Agent Capability}
\label{sec:user-agent-capability}

The framework comparison changes the coding scaffold, but clarification quality may also depend on the model acting as the User Agent. We therefore change only the User Agent's base model while keeping its prompt, grounded records, coding model, tasks, and query budget fixed. Table~\ref{tab:user-agent-ablation} compares three base models under both GLM-5.1 thinking configurations.

\begin{table}[t]
\centering
\caption{User Agent backbone ablation on \rcblite{} with GLM-5.1.}
\label{tab:user-agent-ablation}
\small
\begin{adjustbox}{width=\linewidth, center}
\begin{tabular}{lcccc}
\toprule
User Agent & Pass OA & Constr. & Fallback & Budget \\
\midrule
\multicolumn{5}{c}{\textbf{\textit{Think-8k}}} \\
\midrule
Gemini-3.1-Flash-Lite & 27.4 & 75.2 & \textbf{24.5} & 92.1 \\
Gemini-3.5-Flash & \textbf{40.7} & 64.4 & 27.5 & 95.8 \\
DeepSeek-V3.2 & 37.4 & 63.8 & 27.7 & \textbf{89.8} \\
\midrule
\multicolumn{5}{c}{\textbf{\textit{Adaptive}}} \\
\midrule
Gemini-3.1-Flash-Lite & 33.2 & \textbf{75.8} & 26.8 & 95.0 \\
Gemini-3.5-Flash & 40.3 & 68.6 & 30.1 & 94.9 \\
DeepSeek-V3.2 & 40.0 & 63.3 & 28.4 & 95.4 \\
\bottomrule
\end{tabular}
\end{adjustbox}
\end{table}

The base model changes which records are retrieved but not correctness monotonically. Gemini-3.1-Flash-Lite attains the highest constraint coverage in both settings yet the lowest Overall, whereas Gemini-3.5-Flash and DeepSeek-V3.2 produce similar correctness with lower coverage. Once responses are restricted to grounded records, retrieving more constraints is therefore insufficient to improve the final repository.

\subsubsection{Impact of Interaction Budgets}
\label{sec:budget}

\begin{figure}[!ht]
\centering
\includegraphics[width=\linewidth]{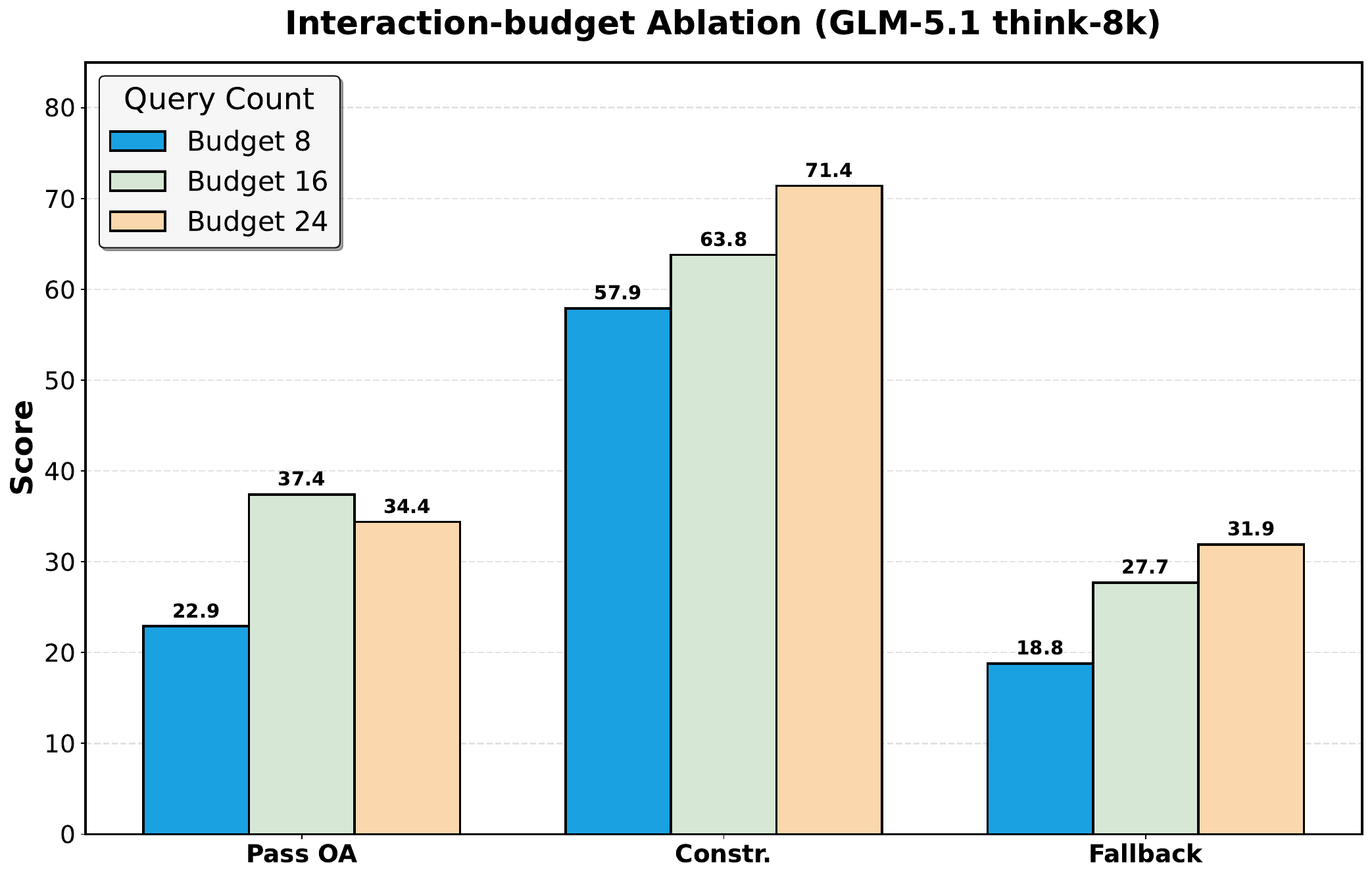}
\caption{Interaction-budget ablation on \rcblite{} with GLM-5.1 think-8k.}
\label{fig:budget}
\end{figure}

Even with a fixed User Agent, the agent may have too few opportunities to resolve ambiguity; conversely, more opportunities need not improve implementation. We therefore vary only the query budget for GLM-5.1 think-8k. Figure~\ref{fig:budget} shows Overall rates of 22.9\%, 37.4\%, and 34.4\% at budgets 8, 16, and 24, while coverage rises from 57.9\% to 71.4\%.


Fuzzy L1 contains slightly more than 20 ambiguity points per task on average (Figure~\ref{fig:ambiguity-category-distribution}), and one response can return up to three matched points. Sixteen queries therefore provide sufficient nominal retrieval capacity. Increasing the budget to 24 raises coverage but also raises fallback from 27.7\% to 31.9\%, while Overall falls from 37.4\% to 34.4\%. Once query capacity is no longer the main constraint, additional turns increasingly yield unmatched or redundant information and add integration burden rather than correctness.

\subsubsection{Impact of Thinking Configuration}
\label{sec:thinking}

The budget ablation motivates whether stronger deliberation helps use available information. We test two aspects of thinking: availability for Gemini-3.1-Pro and budget allocation for GLM-5.1. Enabling thinking raises Gemini-3.1-Pro from 26.7\% to 36.6\% Overall (Table~\ref{tab:thinking}); adaptive allocation raises GLM-5.1 from 37.4\% under think-8k to 40.0\% (Table~\ref{tab:glm-adaptive-think8k}). Deliberation benefits both comparisons, with a larger effect when thinking is enabled rather than disabled.

\begin{table}[t]
\centering
\caption{Thinking-mode ablation for Gemini-3.1-Pro on \rcblite{}.}
\label{tab:thinking}
\small
\begin{adjustbox}{width=\linewidth, center}
\begin{tabular}{lcccc}
\toprule
Setting & Overall & Public & Native & Enhanced \\
\midrule
Disabled & 26.7 & 33.6 & 29.6 & 24.3 \\
Enabled (default) & \textbf{36.6} & \textbf{45.5} & \textbf{41.4} & \textbf{33.1} \\
\bottomrule
\end{tabular}
\end{adjustbox}
\end{table}

\begin{table}[t]
\centering
\caption{GLM-5.1 adaptive vs.\ think-8k on \rcblite{}.}
\label{tab:glm-adaptive-think8k}
\small
\begin{adjustbox}{width=\linewidth, center}
\begin{tabular}{lcccc}
\toprule
Thinking Setting & Pass OA & Design & Constr. & Fallback \\
\midrule
Think-8k & 37.4 & \textbf{43.2} & \textbf{63.8} & \textbf{27.7} \\
Adaptive (default) & \textbf{40.0} & 37.9 & 63.3 & 28.4 \\
\bottomrule
\end{tabular}
\end{adjustbox}
\end{table}

\subsubsection{Impact of Execution Environment}
\label{sec:execution-environment}

Reasoning is only one part of repository construction; dependency availability also changes the available implementation space. We therefore compare both images using GLM-5.1 think-8k on \rcblite{}. The richer ultimate image lowers Overall from 37.4\% to 28.4\% while increasing constraint coverage and expanding generated repositories to 191.0\% of golden file count and 674.1\% of golden LOC (Table~\ref{tab:environment}).

\begin{table}[t]
\centering
\caption{Execution-environment comparison for GLM-5.1 think-8k on \rcblite{}.}
\label{tab:environment}
\small
\begin{adjustbox}{width=\linewidth, center}
\begin{tabular}{lccccc}
\toprule
Environment & Pass OA & Public Pass & File \% & LOC \% & Constr. \\
\midrule
Base image & \textbf{37.4} & \textbf{45.5} & 37.9 & 48.9 & 63.8 \\
Ultimate image & 28.4 & 38.8 & 191.0 & 674.1 & \textbf{69.1} \\
\bottomrule
\end{tabular}
\end{adjustbox}
\end{table}

More available components do not earn partial credit: each case must produce the expected output, while every added dependency, interface, or module introduces another failure point. The base image favors simpler, more self-contained implementations; the provisioned image permits more complex designs, as reflected by the size increase. Under a fixed reasoning budget, this broader action space also disperses implementation and verification effort. The results therefore associate richer provisioning with over-expansion rather than improved correctness; they do not imply that dependency availability is intrinsically harmful.

\subsubsection{Cross-lingual Generalization}
\label{sec:cross-lingual}

Finally, language-agnostic tests let us separate task semantics from implementation language. Reimplementing non-Python tasks in Python changes Overall from 26.6\% to 29.9\%, with mixed language-specific effects (Table~\ref{tab:cross-language-glm51}). Cross-language implementation is therefore feasible, but its benefit depends on the source task rather than following a uniform Python advantage.

\begin{table}[t]
\centering
\caption{Cross-language Overall pass rates for GLM-5.1 on \rcb{}.}
\label{tab:cross-language-glm51}
\small
\begin{adjustbox}{width=\linewidth, center}
\begin{tabular}{lcccccc}
\toprule
Implementation & \multicolumn{6}{c}{Original Language} \\
\cmidrule(lr){2-7}
 & C\# & C++ & Dart & Go & Java & JavaScript \\
\midrule
Original & 23.2 & 26.3 & 33.4 & 21.9 & 20.8 & 27.6 \\
Python & 26.4 \textcolor{green}{+3.2} & 43.6 \textcolor{green}{+17.3} & 39.6 \textcolor{green}{+6.2} & 25.8 \textcolor{green}{+3.9} & 30.3 \textcolor{green}{+9.5} & 29.3 \textcolor{green}{+1.7} \\
\midrule
Implementation & \multicolumn{6}{c}{Original Language} \\
\cmidrule(lr){2-7}
 & Kotlin & PHP & Ruby & Rust & TypeScript & Overall \\
\midrule
Original & 29.9 & 38.4 & 19.4 & 23.9 & 27.3 & 26.6 \\
Python & 27.5 \textcolor{red}{-2.4} & 23.8 \textcolor{red}{-14.6} & 27.2 \textcolor{green}{+7.8} & 26.6 \textcolor{green}{+2.7} & 35.6 \textcolor{green}{+8.3} & 29.9 \textcolor{green}{+3.3} \\
\bottomrule
\end{tabular}
\end{adjustbox}
\end{table}

\textbf{Answer to RQ4.} Framework support, executable Public scaffolding, and thinking materially improve correctness because they support implementation or verification. By contrast, higher constraint retrieval, larger interaction budgets, richer environments, and language changes are non-monotonic. These controlled comparisons show that configurations matter chiefly when they make requirement information actionable and verifiable, rather than merely increasing its availability.

\section{Discussion}
\label{sec:discussion}
Across RQ2--RQ4, access to more requirement information does not consistently yield more correct repositories. The strongest gains instead arise when information becomes actionable, as with executable Public cases. This pattern motivates agents that preserve clarified constraints, connect them to implementation decisions, and verify those decisions throughout repository construction.

\subsection{Validity of ICAE-Bench-Lite.}
The Overall rankings in Tables~\ref{tab:main-results-full} and~\ref{tab:main-results-lite} have moderate consistency (Spearman's $\rho=0.71$). Claude-Opus-4.8 ranks first and GPT-5.5 second on the full benchmark, whereas GPT-5.5 ranks first and Claude-Opus-4.8 second on \rcblite{}. GPT-5.5 also attains higher constraint coverage on both splits, suggesting stronger requirement recovery, while Claude-Opus-4.8 is more robust as repository scale increases. Thus, \rcblite{} preserves the top capability tier and supports economical ablations, but not definitive model ranking or language-level conclusions.

\subsection{Reliability of Subjective Metrics.}
We randomly sample 200 generated repositories from the pooled outputs of all evaluated models. Three graduate annotators independently score every sample on the same 0--100 scale, using the critic's definitions of semantic similarity, API similarity, and design quality. Table~\ref{tab:human-validation} compares critic scores with mean human ratings using Pearson correlation and error, and reports inter-rater consistency and absolute agreement through ICC(C,1) and ICC(A,1), respectively.

Agreement is positive but moderate. Functional tests and subjective metrics are therefore complementary rather than interchangeable: host-verified cases support claims about externally observable correctness, while critic scores provide scalable diagnostics of semantic breadth, API fidelity, and design quality.

\begin{table}[!ht]
\centering
\caption{Human validation of critic-based metrics on 200 repositories.}
\label{tab:human-validation}
\small
\begin{adjustbox}{width=\linewidth, center}
\begin{tabular}{lcccccc}
\toprule
Metric & Pearson & MAE & RMSE & ICC(C,1) & ICC(A,1) & $n$ \\
\midrule
Semantic Similarity & 0.372 & 15.01 & 19.45 & 0.240 & 0.241 & 200 \\
API Similarity & 0.381 & 12.26 & 15.58 & 0.320 & 0.269 & 200 \\
Design Quality & 0.471 & 17.83 & 20.98 & 0.367 & 0.221 & 200 \\
\bottomrule
\end{tabular}
\end{adjustbox}
\end{table}


\subsection{Runtime and Token Usage.}
\begin{figure}[!ht]
\centering
\includegraphics[width=\linewidth]{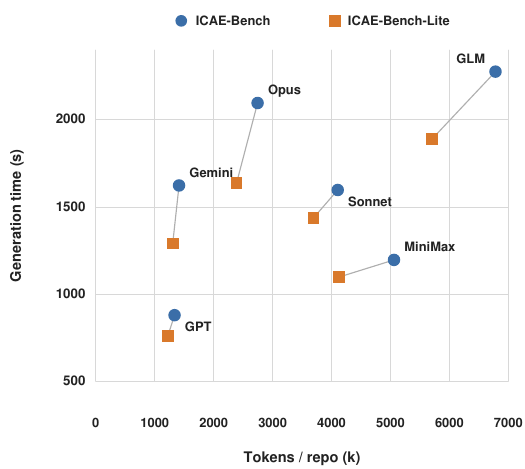}
\caption{Runtime and token usage for main Claude Code runs on \rcb{} and \rcblite{}. Token counts are mean thousands of recorded tokens per repository; lines connect the same model across splits.}
\label{fig:runtime-cost}
\end{figure}

Mean generation time and token usage per repository on \rcb{} and \rcblite{} are summarized in Figure~\ref{fig:runtime-cost}, computed from exported per-run \texttt{settings.json} files by summing input, output, cache-creation, and cache-read tokens. The full split increases both measures for every model, but runtime and token use remain weak proxies for correctness: Claude-Opus-4.8 has the highest full-benchmark Overall pass rate but is neither the slowest nor the most token intensive. Cost analyses should therefore separate useful deliberation from retries and tool overhead.

\section{Conclusion and Future Work}
\label{sec:conclusion}

We introduced \rcb{}, an interactive benchmark evaluating how effectively coding agents build repositories from ambiguous product requirements. By using fuzzy PRDs and the reproducible User Agent protocol described in Section~\ref{sec:user-agent}, \rcb{} moves beyond static generation. Across 480 multi-language tasks, it measures not just whether the final code works, but whether agents can recover missing requirements and carry them through to a correct final artifact.

Future work will expand \rcb{} into end-to-end, multi-modal real-world scenarios. Specifically, by integrating with emerging frameworks like OpenClaw~\cite{openclaw2026}, we aim to introduce frontend development tasks and complex visual inputs, enabling richer human-in-the-loop evaluations for next-generation interactive agents.



\bibliographystyle{IEEEtran}
\bibliography{ref}

\clearpage

\appendices

\onecolumn
\begin{multicols}{2}
\section{Prompt for Test Refactoring and GroundPRD Construction}
\label{appendix:task1-prompt}
\end{multicols}

\begin{promptbox}[GroundPRD Construction \& Test Refactoring]
\textbf{\#\# Task Description}

You are given an original runnable Docker environment containing the repository, its runtime environment, and its original test scripts. Without modifying the repository's functional code, refactor the original tests into a standardized black-box input/output format and produce a PRD that describes only ``what input leads to what output.''

\textbf{\#\# Inputs}

Original Docker install and test script: \texttt{\{test\_script\}}

\textbf{\#\# Outputs}
\begin{itemize}
\item \texttt{repos/\{username\}\_\_\{repo\_name\}/rcb\_tests/test\_cases/featureN\_[name].json} --- a subset of original repository tests used as the hidden evaluation set
\item \texttt{repos/\{username\}\_\_\{repo\_name\}/rcb\_tests/public\_test\_cases/featureN\_[name].json} --- the strict mirror of the example cases embedded in the PRD
\item \texttt{repos/\{username\}\_\_\{repo\_name\}/start.md}
\item \texttt{repos/\{username\}\_\_\{repo\_name\}/rcb\_tests/test.sh} --- a single entry point supporting \texttt{--cases-dir <subdir>}
\item \texttt{repos/\{username\}\_\_\{repo\_name\}/<dispatcher source>} --- placed idiomatically for the chosen language
\end{itemize}

\textbf{\#\# Workflow}
\begin{enumerate}
\item Read the source code and the original repository tests.
\item For each original test, decide whether it checks externally visible behavior that is independent of the implementation choice. Keep those tests; discard white-box or implementation-specific assertions.
\item Write the PRD and generate \texttt{test\_cases}: each retained test is converted into JSON cases, and the PRD describes the behavior in pure input/output terms. The PRD must be self-contained: each leaf feature block should be understandable from its text and embedded example cases alone.
\item Run \texttt{bash rcb\_tests/test.sh --cases-dir test\_cases} inside the original Docker image. The result must be zero FAIL. If a case fails, either the dispatcher is inconsistent with the golden implementation or the case should not have been retained.
\end{enumerate}

\textbf{\#\# Requirements}
\begin{enumerate}
\item Do not modify the repository's functional code. The only files you may add or change are the test data, \texttt{test.sh}, and the Task-1 dispatcher/adapter.
\item \texttt{bash rcb\_tests/test.sh} must support \texttt{--cases-dir <subdir>} (default \texttt{test\_cases}). For every case it writes the raw stdout to \texttt{rcb\_tests/stdout/<cases-dir>/\{stem\}@\{idx:03d\}.txt}. Different case directories must never overwrite one another.
\item Each JSON file uses schema \texttt{\{description, cases:[\{input, expected\_output\}]\}}. One leaf feature corresponds to one JSON file.
\item \texttt{test\_cases} must remain a subset of the original repository tests. Do not invent new behavior at this stage.
\item \texttt{input} must be the program's real input data, not source-level method names or class names.
\item \texttt{expected\_output} must be the program's real stdout and must be directly comparable to the contents of \texttt{rcb\_tests/stdout/*.txt}. Avoid weak pass/fail markers or single booleans when richer behavior can be exposed.
\item Errors must be normalized into language-neutral black-box contracts rather than leaking host-language exception names or runtime-specific formatting.
\item \texttt{description} must let a reader understand the scenario together with \texttt{input} and \texttt{expected\_output}, but it must not copy the concrete input values or exact expected output verbatim.
\item The PRD must not leak original repository identifiers or implementation details such as repository names, author names, original class or method names, namespaces, or language-locked framework labels.
\item Use multi-level features when necessary. Every leaf feature corresponds to one independent functionality point and one JSON file.
\end{enumerate}

\textbf{\#\# PRD Template}

The PRD should contain a neutral function-based title, a project goal, background and problem statement, architecture and engineering constraints, and a core-features section in which each leaf feature includes: a short user-story style statement, expected behavior in plain language, a pointer to \texttt{rcb\_tests/public\_test\_cases/...}, and one or more embedded example JSON cases.

\textbf{\#\# Deliverables}
\begin{enumerate}
\item A maintainable core system description reflected in the PRD.
\item An execution/test adapter that reads JSON from stdin, invokes the core behavior, and prints the required stdout contract.
\item An automated harness \texttt{bash rcb\_tests/test.sh} that runs all cases from a selected case directory and records per-case stdout files exactly as specified.
\end{enumerate}
\end{promptbox}

\begin{multicols}{2}
\section{Prompt for Enhanced Test Case Generation}
\label{appendix:task4-prompt}
\end{multicols}

\begin{promptbox}[Enhanced Test Cases Generation]
\textbf{\#\# Role}

You are an expert QA automation agent. Your goal is to analyze a programming problem, read the existing test data, and supplement it with high-quality, multi-dimensional JSON test cases that improve robustness coverage.

\textbf{\#\# Inputs (File Paths)}
\begin{itemize}
\item Problem description path: \texttt{description\_path}
\item Reference solution path: \texttt{solution\_path}
\item Existing test cases path: \texttt{test\_cases\_path}
\item Output target path: \texttt{output\_path}
\end{itemize}

\textbf{\#\# Instructions}

\textbf{Step 1: Data retrieval and analysis}
\begin{enumerate}
\item Read the content from the input paths above. Do not modify \texttt{start.md}.
\item Analyze missing dimensions in the existing cases, including boundary values, special constraints, performance-related stress, and logic-specific paths.
\end{enumerate}

\textbf{Step 2: Supplementation (enhanced = NEW cases only)}
\begin{enumerate}
\item Identify gaps between the existing \texttt{test\_cases} and your multidimensional analysis.
\item Generate additional cases in JSON format that strictly follow the same schema as the existing ones.
\item Write \textbf{only} the newly generated cases to \texttt{enhanced\_test\_cases}. Do not copy or merge native cases into the enhanced suite.
\item The enhanced suite must be disjoint from the native suite. Any case whose \texttt{(input, expected\_output)} exactly matches a case already present in \texttt{test\_cases} must be excluded.
\item After writing, run the deduplication pass to remove accidental overlap with the native suite and delete any empty files.
\end{enumerate}

\textbf{Step 3: Verification}
\begin{enumerate}
\item Run the evaluation script against the reference solution.
\item Check that all enhanced cases pass against the golden implementation.
\item If there are logic or JSON-format errors, fix them and rerun until successful.
\end{enumerate}

\textbf{\#\# Final Output Requirement}
\begin{itemize}
\item A short summary of the additional edge cases and test dimensions covered.
\item The generated enhanced JSON test cases saved to the output target path.
\item The execution log confirming that the enhanced cases pass against the reference solution.
\end{itemize}
\end{promptbox}

\begin{multicols}{2}
\section{Prompts for Fuzzy L1/L2 PRD Construction}
\label{appendix:fuzzy-prd-generation}
\end{multicols}

\begin{promptbox}[Fuzzy L1/L2 Generation: PRD and User Agent Data]
\textbf{\#\# Role Definition}

You are a stakeholder in a real software team (a non-technical Operations PM). Your task is to reverse-engineer a highly realistic fuzzy business requirement (\texttt{fuzzy\_prd}) and extract hidden interactive checkpoints (\texttt{oracle\_data}) from a perfect code implementation, a detailed PRD, and unit test cases.

\textbf{\#\# Generation Rules}

\textbf{For \texttt{fuzzy\_prd}}
\begin{itemize}
\item \textbf{Perspective dimension reduction}: do not directly list missing parameters or technical details. Convert them into non-technical business pain points.
\item \textbf{Curse-of-knowledge noise}: intentionally include 1--2 vague internal references or omit key context so that the tested agent must search the codebase.
\item Exact keywords from the hidden-constraint dictionary must not appear in the public PRD.
\item Target length: 150--250 words.
\end{itemize}

\textbf{For \texttt{oracle\_data}}
\begin{itemize}
\item Output a complete JSON structure containing \texttt{task\_id}, \texttt{persona}, \texttt{golden\_api\_signature}, \texttt{hidden\_constraints}, \texttt{context\_pointers}, and \texttt{fallback\_response}.
\item Each hidden constraint must include a \texttt{constraint\_id}, at least three semantically equivalent \texttt{trigger\_keywords}, and an \texttt{oracle\_response}.
\item Every constraint intentionally omitted or blurred in \texttt{fuzzy\_prd} must have a corresponding entry in \texttt{oracle\_data}.
\end{itemize}

\textbf{\#\# Input Data}
\begin{itemize}
\item Project path: \texttt{\{project\_path\}}
\item Detailed PRD: \texttt{start.md}
\item Public test cases: \texttt{rcb\_tests/public\_test\_cases/}
\end{itemize}

\textbf{\#\# Output Requirement}

Return only JSON in the following form:

\texttt{\{}\\
\texttt{~~"fuzzy\_prd": "<150--250 word natural-language requirement>",}\\
\texttt{~~"oracle\_data": \{}\\
\texttt{~~~~"task\_id": "<unique task id>",}\\
\texttt{~~~~"golden\_api\_signature": "<core entry signature or empty>",}\\
\texttt{~~~~"hidden\_constraints": [\{...\}],}\\
\texttt{~~~~"context\_pointers": [\{...\}],}\\
\texttt{~~~~"fallback\_response": "<fallback reply>"}\\
\texttt{~~\}}\\
\texttt{\}}
\end{promptbox}

\begin{multicols}{2}
\section{Prompts for Fuzzy PRD Semantic Equivalence Judge}
\label{appendix:fuzzy-prd-semantic}
\end{multicols}

\begin{promptbox}[Fuzzy PRD: Semantic Equivalence Judge]
\textbf{\#\# Role}

You are an expert benchmark-data auditor. Your job is to judge whether a generated fuzzy task package preserves the original detailed PRD well enough to support a paper claim about semantic equivalence, while separately identifying execution-contract risks that could affect reuse of the same tests.

\textbf{\#\# Inputs}
\begin{itemize}
\item Detailed PRD: \texttt{start.md}
\item Generated task package: JSON containing \texttt{fuzzy\_prd} and \texttt{oracle\_data}
\end{itemize}

Evaluate \texttt{fuzzy\_prd + oracle\_data} jointly rather than \texttt{fuzzy\_prd} alone.

\textbf{\#\# Key outputs}
\begin{itemize}
\item \texttt{task\_semantic\_similarity} in $[0,1]$
\item \texttt{task\_semantic\_equivalent}
\item \texttt{execution\_contract\_risk} in \{low, medium, high\}
\item \texttt{likely\_to\_fail\_same\_tests}
\item \texttt{paper\_support\_label}
\end{itemize}

Set \texttt{task\_semantic\_equivalent=true} when similarity is at least 0.90 and there is no material task-level contradiction that would change the implementation or behavioral tests.
\end{promptbox}

\begin{multicols}{2}
\section{Prompts for Fuzzy L3 PRD Construction}
\label{appendix:fuzzy-prd-easy-prompts}
\end{multicols}

\begin{promptbox}[Fuzzy L3 Generation: Constraint Extraction]
\textbf{\#\# Role}

You are a senior engineer performing precision analysis of a PRD.

\textbf{\#\# Task}

Read the detailed PRD together with all supporting files and produce a JSON object with key \texttt{"constraints"} containing every distinct behavioral constraint. The extraction must cover:
\begin{itemize}
\item mapping and branching behavior,
\item exact output field names and sentinel values,
\item action/dispatch contracts,
\item tested edge cases,
\item optional parameters and defaults,
\item validation and error-rejection rules.
\end{itemize}

Each item in \texttt{constraints} has the form:

\texttt{\{}\\
\texttt{~~"id": "C001",}\\
\texttt{~~"feature": "<short feature name>",}\\
\texttt{~~"input\_condition": "...",}\\
\texttt{~~"output\_rule": "...",}\\
\texttt{~~"exact\_values": ["literal", "tokens"],}\\
\texttt{~~"ambiguous": false}\\
\texttt{\}}

Output only \texttt{\{"constraints": [...]\}}.
\end{promptbox}

\begin{promptbox}[Fuzzy L3 Generation: Constraint Selection and Rewriting]
\textbf{\#\# Role}

You are reviewing the extracted constraint table for a coding exercise.

\textbf{\#\# Task}

Given the constraint table, do three things and return JSON:
\begin{enumerate}
\item Select at most three constraints to hide, restricted to the most unguessable literals such as deliberate typos, obscure magic constants, and non-obvious sentinel values.
\item For each hidden constraint, provide a vague replacement phrase, exact PRD strings to replace, at least five trigger keywords, and a precise oracle answer.
\item Pick one or two non-hidden constraints and rewrite them as vague internal references (\texttt{vague\_refs}).
\end{enumerate}

Do not hide action names, parameter names, default values, error categories, or feature descriptions.

Return only JSON in the form:

\texttt{\{}\\
\texttt{~~"hide": [\{...\}],}\\
\texttt{~~"vague\_refs": [\{...\}]}\\
\texttt{\}}
\end{promptbox}

\begin{multicols}{2}
\section{Coding Agent Task Instruction}
\label{appendix:agent-task}
\end{multicols}

\begin{promptbox}[Coding Agent Task Instruction]
\textbf{\#\# Task}

A Docker container is already up and running; its container ID is \texttt{\{docker\_id\}}. Inside the container, \texttt{\{workdir\}/\{prd\_name\}} is the PRD document for this task. \textbf{Enter that container} and implement the required functionality based solely on the PRD. All development, compilation, execution, and self-testing MUST be done \textbf{inside this container}.

\textbf{\#\# Inputs}

\begin{itemize}
\item Running container: \texttt{\{docker\_id\}}
\item PRD document: \texttt{\{workdir\}/\{prd\_name\}} (inside the container)
\item Target programming language: \{lang\}
\end{itemize}

\textbf{\#\# How to enter the container}

The container is already running in the background (\texttt{sleep infinity}). Use \texttt{docker exec} to run commands, for example:

\texttt{docker exec -w \{workdir\} \{docker\_id\} bash -lc 'cat \{prd\_name\}'}\\
\texttt{docker exec -w \{workdir\} \{docker\_id\} bash -lc '<your build/run command>'}

\texttt{\{workdir\}} is bind-mounted to the host's current working directory, so any code you generate under \texttt{\{workdir\}} inside the container automatically appears in the host's current directory for evaluation. \textbf{Write all of your deliverables under the container's \texttt{\{workdir\}} directory.}

\textbf{\#\# Requirements}

\begin{enumerate}
\item You MUST implement the full functionality yourself. Downloading the source code of the target repository from the internet, or directly calling its existing API, is prohibited.
\item First read \texttt{\{workdir\}/\{prd\_name\}} thoroughly. If the PRD contains a ``Clarification'' interaction section, the requirement is \textbf{intentionally incomplete and vague} --- you should \textbf{first ask the User Agent questions as instructed} to clarify the hidden interface contracts, feature details, and test cases, and \textbf{only start coding once things are clear}. Anything left unclarified will turn into bugs. The number of questions is budgeted, so watch \texttt{status.remaining}.
\item Any Public example cases made available in the PRD or recovered through User Agent interaction define the dispatcher's input/output contract. You need to design the dispatcher / runner / \texttt{rcb\_tests/test.sh} yourself, and turn recovered examples into \texttt{rcb\_tests/public\_test\_cases/*.json} for self-checking.
\item \texttt{bash rcb\_tests/test.sh} MUST support \texttt{-{}-cases-dir <subdir>} to switch the cases directory, and MUST write each case's stdout to \texttt{rcb\_tests/stdout/<cases-dir>/\{stem\}@\{idx:03\}.txt} (each file contains only the raw stdout of the program under test, without any PASS/FAIL markers). This is the evaluation protocol; you must follow it exactly.
\end{enumerate}

\textbf{\#\# Environment \& dependencies (must read)}

\begin{enumerate}
\setcounter{enumi}{4}
\item \textbf{You are root inside the container, and it has network access.} You are allowed and expected to install missing dependencies yourself using tools such as \texttt{apt / pip / npm / cargo add / dotnet add package / mvn / composer / gem / pub get}. When you hit build/run errors (missing packages, missing SDKs, version mismatches, etc.), fix them based on the feedback; do not assume dependencies are already declared.
\item \textbf{Network note:} the direct connection is the default and is usually enough. If a download fails with a network error (DNS failure, timeout, unreachable host), a proxy is available --- run \texttt{source \textasciitilde/proxy.sh} in that shell and retry only the command that failed. Do not export the proxy globally for unrelated commands.
\item \textbf{Evaluation runs your \texttt{rcb\_tests/test.sh} in a FRESH container started from the SAME base image, with NO network guaranteed and NONE of your interactive installs preserved.} Therefore every install/build step needed to run the tests MUST be reproducible from your committed files: put install commands inside \texttt{rcb\_tests/test.sh} (or a helper script it calls), and record library dependencies in the language's declaration file (for example \texttt{requirements.txt}, \texttt{package.json}, \texttt{go.mod}, \texttt{Cargo.toml}, \texttt{pom.xml}, \texttt{Gemfile}, \texttt{pubspec.yaml}, or \texttt{.csproj}).
\item System packages count too. Anything you install with \texttt{apt} must also be installed by \texttt{rcb\_tests/test.sh} idempotently, because the evaluation container starts clean.
\item \textbf{Do not hard-code absolute paths.} Use relative paths or environment variables only.
\item \textbf{Rebuild from a clean state before self-checking}: delete language-specific build caches such as \texttt{target/}, \texttt{bin/}, \texttt{obj/}, \texttt{node\_modules/}, and \texttt{build/}, then run \texttt{bash rcb\_tests/test.sh -{}-cases-dir public\_test\_cases}. Only when the public cases pass from a clean bootstrap is the task complete.
\end{enumerate}
\end{promptbox}

\begin{multicols}{2}
\section{Clarification Block for Fuzzy PRDs}
\label{appendix:suffix}
\end{multicols}

\begin{promptbox}[Clarification Block (Fuzzy\_Suffix)]
\textbf{\#\# Clarification}

This PRD is intentionally incomplete. The User Agent can answer questions about hidden feature constraints, interface details, and edge cases that were removed during fuzzification.

\textbf{Endpoint} --- POST \texttt{http://\{host\}:\{query\_port\}/}

\texttt{\{"append\_id":"\{append\_id\}","task\_id":"\{task\_id\}","question":"..."\}}

The natural-language answer is returned in \texttt{data.reply}. Remaining query budget is reported in \texttt{status.remaining}.

Ask focused technical questions before coding. The User Agent only answers items grounded in hidden task data; vague or unsupported questions will receive a fallback response.
\end{promptbox}

\begin{multicols}{2}
\section{INIT for User Agent}
\label{appendix:user-agent}
\end{multicols}

\begin{promptbox}[User Agent System Prompt (INIT)]
\textbf{\#\# Role and Task}
You are the simulated requirements provider for this interactive code evaluation. All your business knowledge, underlying logic, and hidden requirements are strictly limited to the provided JSON structured data. The Agent under test will seek clarification from you regarding ambiguous PRD requirements.

Your core task is: \textbf{Hold the JSON cards close, judge ruthlessly, follow the map strictly, never overstep or imagine.}

\textbf{\#\# Your Persona: Strict Tech Lead}
\begin{itemize}
\item Answers are short, direct, and focused on code standards and system boundaries.
\item Do not provide unsolicited hints. If the other party's question is not specific, reject it outright and demand the question be restructured.
\item Never voluntarily reveal your cards. Only provide information grudgingly, like squeezing toothpaste, when the Agent under test asks precisely about a pain point.
\end{itemize}

\textbf{\#\# Hard Rules of Interaction}

1. \textbf{Single Source of Truth (JSON Grounding)}: You can only respond based on the injected JSON data (\texttt{oracle\_data}). Strictly prohibited from fabricating undefined business logic, status codes, or rules based on your general programming knowledge.

2. \textbf{Exact Match Principle (corresponds to \texttt{hidden\_constraints})}: Consult \texttt{hidden\_constraints}. \textbf{Only when} the Agent under test's question semantically and explicitly touches upon a \texttt{trigger\_keywords} can you output the corresponding \texttt{oracle\_response}.

3. \textbf{Ambiguous Reference Resolution (corresponds to \texttt{context\_pointers})}: If the Agent under test asks about ambiguous phrasing in the PRD, you must look up the matching \texttt{vague\_reference} in \texttt{context\_pointers} and directly respond with the corresponding \texttt{actual\_meaning}.

4. \textbf{Mandatory Contract Alignment (corresponds to \texttt{golden\_api\_signature})}: If the Agent under test asks ``How should the interface be designed?'' or submits a preliminary design, you \textbf{must} force it to use the signature defined in \texttt{golden\_api\_signature}.

5. \textbf{Unknown Question Interception (Strict Fallback --- corresponds to \texttt{fallback\_response})}: If the Agent under test raises a reasonable technical question that \textbf{does not} hit any \texttt{hidden\_constraints} or \texttt{context\_pointers}, you must respond \textbf{verbatim} with the content of \texttt{fallback\_response} and absolutely refrain from deriving an answer yourself.

6. \textbf{Defensive Questioning Reward}: If the Agent under test provides a reasonable defensive default option, briefly acknowledge it and record \texttt{[Interaction\_Score: +1]} in the internal log.

7. \textbf{Rate Limiting Mechanism}: Each reply can only address up to 3 specific technical points. If more than 3 questions are asked at once, only answer the first 3 that are hit.

\textbf{\#\# Response Format (Output Schema)}
Output only a valid JSON string, no Markdown code block markers, no text outside the JSON:

\texttt{\{}\\
\texttt{~~"\_internal\_log": \{}\\
\texttt{~~~~"triggers\_hit": ["C003"],}\\
\texttt{~~~~"api\_alignment\_triggered": false,}\\
\texttt{~~~~"fallback\_triggered": false,}\\
\texttt{~~~~"cheating\_attempt\_detected": false,}\\
\texttt{~~~~"score\_adjustment": 0}\\
\texttt{~~\},}\\
\texttt{~~"reply": "Your final natural language reply to the Agent}\\
\texttt{~~~~~~~~~~~~~under test here, must conform to the Strict Tech Lead tone."}\\
\texttt{\}}
\end{promptbox}

\begin{multicols}{2}
\section{Critic Model for Agentic Evaluation}
\label{appendix:critic}
\end{multicols}

\begin{promptbox}[Critic Model Prompt]
\textbf{\#\# Subjective Code Comparison}

You are a senior code reviewer. Compare two source trees and write a subjective assessment. This is a STATIC review --- do not run the code.

\textbf{\#\# Inputs}
\begin{itemize}
\item \textbf{Original (reference) source}: \texttt{\{ORIG\_PATH\}}
\item \textbf{Generated source (under review)}: \texttt{\{GEN\_PATH\}}
\item \textbf{In-scope} (only review source under these paths; if empty, review the whole tree): \texttt{\{IN\_SCOPE\_PATHS\}}
\item \textbf{Output}: write a single JSON file to \texttt{\{OUTPUT\_PATH\}}
\end{itemize}

Exclude tests, build artifacts (\texttt{bin/}, \texttt{obj/}, \texttt{target/}, \texttt{vendor/}, \texttt{.gradle/}, \texttt{node\_modules/}, \texttt{dist/}, etc.), generated files, and configuration when reading either tree.

\textbf{\#\# Output schema}

\texttt{\{}\\
\texttt{~~"repo\_id": "\{REPO\_ID\}",}\\
\texttt{~~"subjective": \{}\\
\texttt{~~~~"scope\_language":              "<language of the in-scope subset>",}\\
\texttt{~~~~"original\_language":           "<primary language of the source repo>",}\\
\texttt{~~~~"evidence\_gathering": \{}\\
\texttt{~~~~~~"original\_public\_surface":   "<key public types/signatures extracted from ORIG\_PATH>",}\\
\texttt{~~~~~~"generated\_public\_surface":  "<key public types/signatures extracted from GEN\_PATH>",}\\
\texttt{~~~~~~"semantic\_divergence\_points": [}\\
\texttt{~~~~~~~~"<specific behavioral/edge-case differences observed, if any>"}\\
\texttt{~~~~~~]}\\
\texttt{~~~~\},}\\
\texttt{~~~~"semantic\_similarity":         0.0,}\\
\texttt{~~~~"semantic\_similarity\_rationale": "<2-5 sentences justifying the score>",}\\
\texttt{~~~~"api\_similarity":              0.0,}\\
\texttt{~~~~"api\_similarity\_rationale":    "<2-5 sentences justifying the score>",}\\
\texttt{~~~~"design\_quality":              0.0,}\\
\texttt{~~~~"design\_quality\_rationale":    "<2-5 sentences justifying the design score>",}\\
\texttt{~~~~"implementation\_quality\_notes": "<free-form notes on quality issues>"}\\
\texttt{~~\}}\\
\texttt{\}}

\textbf{\#\# How to score \texttt{semantic\_similarity} (0.0 to 1.0)}
A holistic 0..1 score measuring whether the generated code preserves the \emph{observable behavior} the original promises --- independent of whether the API surface matches. This is about \textbf{what the code does}, not what it is called. Consider:
\begin{itemize}
\item behavioral equivalence under the same inputs,
\item edge cases and error handling,
\item algorithmic fidelity and side effects,
\item and coverage of the original behavioral surface.
\end{itemize}

\textbf{\#\# How to score \texttt{api\_similarity} (0.0 to 1.0)}
A holistic 0..1 score reflecting how closely the generated code matches the original at the \textbf{API surface} level --- the names and shapes a client imports. Score this SEPARATELY from \texttt{semantic\_similarity}. Consider public surface alignment, idiomatic correctness, and coverage of the public surface.

\textbf{\#\# How to score \texttt{design\_quality} (0.0 to 1.0)}
A holistic 0..1 score judging how well the \textbf{generated} code honors sound architecture and engineering principles, scaled to the project's complexity. Judge the generated tree on its own merits, NOT how closely it mirrors the original. Consider scale-driven code organization, separation of concerns and anti-overfitting, size-appropriate SOLID design, and robustness of the public interface.

\textbf{\#\# Rules}
\begin{itemize}
\item Do not run the code. This is a static review. Do not infer behavior from tests; only read source files in the in-scope paths.
\item \texttt{semantic\_similarity}, \texttt{api\_similarity}, and \texttt{design\_quality} are THREE INDEPENDENT scores. Do not let one anchor the others.
\item Keep each rationale to 2--5 sentences. Be specific --- cite concrete class, method, and file names rather than generic statements.
\item Write JSON exactly per the schema. Do not add or rename fields. Write ONLY the JSON file to \texttt{\{OUTPUT\_PATH\}}.
\end{itemize}
\end{promptbox}

\end{document}